\def\year{2021}\relax
\documentclass[letterpaper]{article} 
\usepackage{aaai21}  
\usepackage{times}  
\usepackage{helvet} 
\usepackage{courier}  
\usepackage[hyphens]{url}  
\usepackage{graphicx} 
\usepackage{natbib}  
\usepackage{caption} 
\usepackage[utf8]{inputenc} 
\usepackage[T1]{fontenc}    
\usepackage{url}            
\usepackage{booktabs}       
\usepackage{amsfonts}       
\usepackage{nicefrac}       
\usepackage{microtype}      
\usepackage{color,colortbl}
\usepackage{mwe}
\usepackage{subcaption} 
\usepackage{natbib}
\usepackage{appendix}
\usepackage{dirtytalk}
\usepackage{amsfonts}
\usepackage{graphicx}
\usepackage{amsmath}
\usepackage{amssymb}
\usepackage{mathtools}
\usepackage{placeins}
\usepackage{siunitx}
\usepackage{verbatim}
\usepackage{caption} 
\usepackage{longtable}
\usepackage{placeins}
\usepackage{bbm}
\usepackage{pdfpages}
\usepackage[skip=2pt]{caption}
\usepackage[capitalise]{cleveref}
\newcommand{\eg}{e.g.\xspace}
\newcommand{\ie}{i.e.\xspace}

\newcommand{\iid}{i.i.d.\xspace}
\newcommand{\niid}{non-i.i.d.\xspace}
\newcommand{\ts}{\textsuperscript}

\DeclareMathOperator*{\argmin}{arg\,min}
\DeclareMathOperator*{\g}{\textbf{g}}
\DeclareMathOperator{\sign}{sgn}

\newcommand{\IND}{{I_{\left | \sum_{k \in S_t} \sign(\Delta_{t,i}^k) \right |\geq \theta}}}

\title{Defending against Backdoors in Federated Learning \\ with Robust Learning Rate}
\author{
    Mustafa Safa Ozdayi,
    Murat Kantarcioglu, 
    Yulia R. Gel \\
}
\affiliations{
    \textsuperscript{\rm}The University of Texas at Dallas \\
    
    \{mustafa.ozdayi, muratk, ygl\}@utdallas.edu

}

\begin{document}
\maketitle

\begin{abstract}
Federated learning (FL) allows a set of agents to collaboratively train a model without sharing their potentially sensitive data. This makes FL suitable for privacy-preserving applications. At the same time, FL is susceptible to adversarial attacks due to decentralized and unvetted data.
One important line of attacks against FL is the backdoor attacks. In a backdoor attack, an adversary tries to embed a backdoor functionality to the model during training that can later be activated to cause a desired misclassification. To prevent backdoor attacks, we propose a lightweight defense that requires minimal change to the FL protocol. At a high level, our defense is based on carefully adjusting the aggregation server's learning rate, \emph{per dimension} and \emph{per round}, based on the sign information of agents' updates. We first conjecture the necessary steps to carry a successful backdoor attack in FL setting, and then, explicitly formulate the defense based on our conjecture. Through experiments, we provide empirical evidence that supports our conjecture, and we test our defense against backdoor attacks under different settings. We observe that either backdoor is completely eliminated, or its accuracy is significantly reduced. Overall, our experiments suggest that our defense significantly outperforms some of the recently proposed defenses in the literature. We achieve this by having minimal influence over the accuracy of the trained models. In addition, we also provide convergence rate analysis for our proposed scheme.
\end{abstract}

\section{Introduction}
\emph{Federated learning} (FL)~\cite{fed-learning:google} has been introduced as a distributed machine learning protocol. Through FL, a set of agents can collaboratively train a model without sharing their data with each other, or any other third party. This makes FL suitable to settings where data privacy is desired. 
In this regard, FL differs from the traditional distributed learning setting in which data is first centralized at a place, and then distributed to the agents~\cite{dean2012large,li2014scaling}. 
 
At the same time, FL has been shown to be susceptible to \emph{backdoor attacks}~\cite{arxiv:2018:fedlens,arxiv:2018:backdoor}. In a backdoor attack, an adversary disturbs the training process to make the model learn a \emph{targeted misclassification functionality}~\cite{chen2017targeted,shafahi2018poison,liu2017trojaning}. 
In centralized setting, this is typically done by \emph{data poisoning}. For example, in a classification task involving cars and planes, the adversary could label all blue cars in the training data as plane in an attempt to make the model to classify blue cars as plane at the inference/test phase. 
In FL, since the data is decentralized, it is unlikely that an adversary could access all the training data. Thus, backdoor attacks are typically carried through \emph{model poisoning} in the FL context ~\cite{arxiv:2018:fedlens,arxiv:2018:backdoor,sun2019really}. That is, the adversary tries to construct a malicious update that encodes the backdoor in a way such that, when it is aggregated with other agents' updates, the aggregated model exhibits the backdoor. 

In this work, we study backdoor attacks against deep neural networks in FL setting, and formulate a defense. Our solution is based on carefully adjusting the learning rate of the aggregation server during the training. Through experiments, we illustrate that our defense can deter backdoor attacks significantly. Further, this achieved with only minimal degradation in the trained model's accuracy in both \iid and \niid settings. We provide empirical evidence justifying the effectiveness of our defense, and also theoretically analyze its convergence properties. 
In summary, our work significantly outperforms some of the existing defenses in the literature, and succeeds in scenarios where they fail.

The rest of the paper is organized as follows. In Section~\ref{sec:background}, we provide the necessary background information. In Section~\ref{sec:robustLR}, we explain our defense, and in Section~\ref{sec:exps}, we illustrate the performance of our defense under  different experimental settings. In Section~\ref{sec:discussion}, we discuss and elaborate upon the results of our experiments, and finally, in Section~\ref{sec:conc}, we provide a few concluding remarks.
\section{Background}\label{sec:background}

\paragraph{Federated Learning (FL)}
At a high level, FL  is multi-round protocol between an aggregation server and a set of agents in which agents jointly train a model. Formally, participating agents try to minimize the average of their loss functions
$$
\argmin_{w \in R^d} f(w) = \frac{1}{K}\sum_{k=1}^K f_k(w),
$$
where $f_k$ is the loss function of k\ts{th} agent. For example, for neural networks, $f_k$ is typically empirical risk minimization under a loss function $L$ such as cross-entropy, \ie,

$$
f_k(w) = \frac{1}{n_k} \sum_{j=1}^{n_k} L(x_j, y_j; w),
$$
with $n_k$ being the total number of samples in agent's dataset and $(x_j,y_j)$ being the j\ts{th} sample.

Concretely, FL protocol is executed as follows: at round $t$, server samples a subset of agents $S_t$, and sends them $w_t$, the model weights for the current round. Upon receiving $w_t$,  k\ts{th} agent initializes his model with the received weight, and trains for some number of iterations, \eg, via stochastic gradient descent (SGD), and ends up with weights $w_t^k$. The agent then computes his update as $\Delta_t^k = w_t^k - w_t$, and sends it back to the server. Upon receiving the update of every agent in $S_t$, server computes the weights for the next round by aggregating the updates with an aggregation function $\mathbf{g} \colon R^{|S_t| \times d } \rightarrow R^d$ and adding the result to $w_t$. That is,
$w_{t+1} = w_t + \eta\cdot\g( \{ \Delta_t \} )$ where $\{ \Delta_t \} = \cup_{k \in S_t}\Delta_t^k$, and $\eta$ is the server's learning rate.
For example, original FL paper~\cite{fed-learning:google} and many subsequent papers on FL~\cite{arxiv:2018:fedlens,arxiv:2018:backdoor,sun2019really,bonawitz2016practical,geyer2017differentially} consider weighted averaging to aggregate updates. In this context, this aggregation is referred as Federated Averaging (FedAvg), and yields the following update rule,
\begin{equation}
w_{t+1} = w_t + \eta \frac{\sum_{k \in S_t} n_k \cdot \Delta_t^k}{\sum_{k \in S_t} n_k}.
\label{eqn:fedavg}
\end{equation}
In practice, rounds can go on indefinitely, as new agents can keep joining the protocol, or until the model reaches some desired performance  metric (\eg, accuracy) on a validation dataset maintained by the server. 

\paragraph{Backdoor Attacks and Model Poisoning}
\label{sec:bd_attacks}
Training time attacks against machine learning models can roughly be classified into two categories: targeted~\cite{arxiv:2018:fedlens,arxiv:2018:backdoor, chen2017targeted,liu2017trojaning}, and untargeted attacks~\cite{blanchard2017machine,bernstein2018signsgd}.
In untargeted attacks, the adversarial task is to make the model converge to a sub-optimal minima or to make the model completely diverge. Such attacks have been also referred as \emph{convergence attacks}, and to some extend, they are easily detectable by observing the model's accuracy on a validation data. 

On the other hand, in targeted attacks, adversary wants the model to misclassify only a set of chosen samples with minimally affecting its performance on the main task. Such targeted attacks are also known as \emph{backdoor attacks}. A prominent way of carrying backdoor attacks is through \emph{trojans}~\cite{chen2017targeted,liu2017trojaning}. A trojan is a carefully crafted pattern that is leveraged to cause the desired misclassification. For example, consider a classification task over cars and planes and let the adversarial task be making the model classify blue cars as plane. Then, adversary could craft a brand logo, put it on \emph{some} of the blue car samples in the training dataset, and only mislabel those as plane. Then, potentially, model would learn to classify blue cars with the brand logo as plane. At the inference time, adversary can present a blue car sample with the logo to the model to activate the backdoor. Ideally, since the model would behave correctly on blue cars that do not have the trojan, it would not be possible to detect the backdoor on a clean validation dataset.

In FL, the training data is decentralized and the aggregation server is only exposed to model updates. Given that, backdoor attacks are typically carried by constructing malicious updates. That is, adversary tries to create an update that encodes the backdoor in a way such that, when it is aggregated with other updates, the aggregated model exhibits the backdoor. This has been referred as \emph{model poisoning} attack  \cite{arxiv:2018:fedlens,arxiv:2018:backdoor,sun2019really}. For example, an adversary could control some of the participating agents in a FL instance and train their local models on trojaned datasets to construct malicious updates.

\paragraph{Robust Aggregation Methods}
Several works have explored using techniques from robust statistics to deter attacks in FL. At a high level, these works tried replacing averaging with robust estimators\footnote{Informally, a statistical estimator is said to be robust if it cannot be skewed arbitrarily in presence of outliers~\cite{huber19721972}.} such as coordinate-wise median, geometric median, $\alpha$-trimmed mean, or a variant/combination of such techniques~\cite{yin2018byzantinerobust,pillutla2019robust,blanchard2017machine,mhamdi2018hidden}. However, to the best of our knowledge, the primary aim of these defenses are to deter convergence attacks. 

In contrast, a recent work~\cite{sun2019really} has shown FedAvg can be made robust against backdoor attacks in some settings when it is coupled with weight-clipping and noise addition as introduced in~\cite{geyer2017differentially}. Concretely, server inspects updates, and if the $L_2$ norm of an update exceeds a threshold $M$, server clips the update by dividing it with an appropriate scalar. Server then aggregates clipped updates and adds Gaussian noise to the aggregation. 
In this case, the update rule can be written as,
\begin{equation}
\resizebox{0.9\hsize}{!}{$%
w_{t+1} = w_t + \eta 
\bigg( 
\dfrac{\sum_{k \in S_t} n_k \cdot \frac{\Delta_t^k}{max(1, \Vert\Delta_t^k\Vert_2/M)}}{\sum_{k \in S_t} n_k} + \mathcal{N}(0, \sigma^2M^2)
\bigg).
$%
}%
\label{eqn:fedavg_DP}
\end{equation}

Another recent work~\cite{fgold} tries to make FL robust by introducing a per-client learning rate rather than having a single learning rate at the server side, yielding the following update rule,
\begin{equation}
w_{t+1} = w_t + \ \frac{\sum_{k \in S_t} \alpha^t_k \cdot n_k  \cdot \Delta_t^k}{\sum_{k \in S_t} n_k}.
\label{eqn:fgoold}
\end{equation}
where $\alpha^t_k \in [0, 1]$ is the k\ts{th} agent's learning rate for the t\ts{th} round.
The exact details of how learning rates are computed can be found in Algorithm 1 of the respective paper. Though, at a high level, the algorithm tries to assign lower learning rates to updates whose directions are similar, as given by cosine similarity. The rationale of this defense is that, assuming adversary's agents share the common backdoor task, their updates will be more similar among themselves than honest updates. Under this assumption, the algorithm will assign lower learning rates to malicious updates, and reduce their effects.
For example, if there are two identical updates, the algorithm assigns 0 as learning rate to both updates.
However, as we observe experimentally in Section~\ref{sec:exps}, their assumption does not hold in some realistic settings for FL. That is, if local data distributions of honest agents exhibit some similarity, algorithm cannot distinguish the adversarial agents and end up assigning everyone either the same, or very similar learning rates throughout the training process.

Finally in~\cite{bernstein2018signsgd}, authors develop a communication efficient, distributed SGD protocol in which agents only communicate the signs of their gradients. In this case, server aggregates the received signs and returns the sign of aggregation to the agents who locally update their models using it. We refer their aggregation technique as \emph{sign aggregation}, and in FL setting, it yields the following update rule,
\begin{equation}
w_{t+1} = w_t + \eta \big( \sign \sum_{k \in S_t} \sign(\Delta_t^k)  \big),
\label{eqn:sign_agg}
\end{equation}
where $\sign$ is the element-wise sign operation.
Although authors show their approach is robust against certain adversaries who carry convergence attacks, \eg, by sending random signs, or by negating the signs of their gradients, in Section~\ref{sec:exps}, we show that it is susceptible against backdoors attacks.
\section{Robust Learning Rate}\label{sec:robustLR}
\paragraph{Backdoor Task vs Main Task}
Let $\Delta_{adv}, \Delta_{hon}$, be the aggregated updates of adversarial, and honest agents respectively. Ideally, $\Delta_{adv}$ should steer the parameters of the model to  $w_{adv}$, which ideally minimizes the loss on both the main, and the backdoor attack task.
At the same time, $\Delta_{hon}$ would want to move the model parameters towards $w_{hon}$ that only minimizes the loss on main task. Our main conjecture is that, assuming $w_{hon}$ and $w_{adv}$ are different points, $\Delta_{adv}$ and   $\Delta_{hon}$ will most likely differ in the directions they specify at least for some dimensions. As we show next, assuming a bound on the number of adversarial agents, we can ensure the model moves away from $w_{adv}$, and moves toward $w_{hon}$, by tuning the server's learning rate based on sign information of updates.

\paragraph{Robust learning rate (RLR)}
Following the above insight, we construct a defense which we denote as \emph{robust learning rate} (RLR) by extending the approach proposed in~\cite{bernstein2018signsgd}. In order to move the model towards a particular direction, for each dimension, we require a sufficient number of votes, in form of signs of the updates. Concretely, we introduce a hyperparameter called \emph{learning threshold} $\theta$ at the server-side. For every dimension where the sum of signs of updates is less than $\theta$, the learning rate is multiplied by -1. This is  \emph{to maximize the loss on that dimension rather than minimizing it}. That is, with a learning threshold of $\theta$, the learning rate for the i\ts{th} dimension is given by,
\begin{eqnarray}
\label{majority}
\eta_{\theta, i} =  
\begin{cases}
\eta & \left | \sum_{k \in S_t} \sign(\Delta_{t,i}^k) \right | \geq \theta, \\
-\eta  & \text{otherwise.}
\end{cases}
\label{eqn:rlrTheta}
\end{eqnarray}

For example, consider FedAvg and let $\eta_\theta$ denote the learning rate vector over all dimensions, \ie, $[\eta_{\theta,1}, \eta_{\theta,2}, \dots, \eta_{\theta,d}]^\top$. Then, the update rule with the robust learning rate takes the form,
\begin{equation}
w_{t+1} = w_t + 
\eta_\theta \odot
\frac{\sum_{k \in S_t} n_k \cdot \Delta_t^k}{\sum_{k \in S_t} n_k},
\label{eqn:fedavg_robust}
\end{equation}
where $\odot$ is the element-wise product operation.
Note that, since we only adjust the learning rate, the approach is agnostic to the aggregation function. For example, we can trivially combine it with update clipping and noise addition as in Equation+~(\ref{eqn:fedavg_DP}).

To illustrate how this might help to maximize adversary's loss, we consider a simple example where the local training consists of a single epoch of full-batch gradient descent. In this case, update of k\ts{th} agent is just the negative of his gradients, \ie, $\Delta_t^k = w_t^k - w_t = (w_t - \nabla f_k(w_t)) - w_t = - \nabla f_k(w_t)$. Then, aggregated updates is just the average of negative of agents' gradients, i.e., $-g_{avg}$. Therefore, if sum of the signs at a dimension $i$ is below $\theta$, that dimension is updated as $w_{t,i} = w_{t,i} + \eta\cdot g_{avg,i}$. Otherwise, it is updated as $w_{t,i} = w_{t,i} -  \eta\cdot g_{avg,i}$. So we see that, for dimensions where the sum of signs is below $\theta$, we are moving towards the direction of gradient, and hence, attempting to maximize loss. For other dimensions, we are moving towards the negative of gradient and attempting to minimize the loss as usual. Therefore, assuming number of adversarial agents is sufficiently below $\theta$, the model would try to move away from $w_{adv}$, and would try to move towards $w_{hon}$.

\paragraph{Convergence Rate}
We now turn to deriving the convergence rate for full-batch FedAvg with RLR. 
Let $f_k(w)=\mathbb{E}_{D_k}[f_k(w, \xi_k)]$ be the loss function of k\ts{th} agent, where $D_k$ is its distribution\footnote{Note that $D_i$ and $D_j$ are not necessarily identical for two different agents $i$ and $j$} and $\xi_k$ is randomness caused by the local batch variability. We use $\mathbb{E}$ to denote expectation in respect to all random variables.  Let $g_k$ be the gradient of the k\ts{th} agent at the t\ts{th} rounds, i.e. $g^t_k=\nabla f_k(w^k_{t-1}, \xi^t_k)$,
and $\mathbb{E}_{D_k} (g^t_k|\digamma_t)=\nabla f_k(w^k_{t-1})$ where
$\digamma_t$ is a filtration
generated by all random variables at step $t$, i.e. a sequence of increasing $\sigma$-algebras $\digamma_s \subseteq \digamma_t$ for all $s<t$.
{
Finally, following~\citet{bernstein2018signsgd}, we assume that for all $t, k\in \mathbb{Z}$
each component of the
stochastic gradient vector $g^t_k$ has a unimodal distribution that satisfies population weighted symmetry~\cite{wolfe1974characterization}. In particular, let $W$ be 
a random variable symmetric around zero, i.e., $Pr (W \leq -w) = Pr(W \geq w)$ for each $w>0$. We now consider a family of asymmetric distributions which are constructed by distorting an arbitrary symmetric distribution with a scalar parameter $\beta>0$ such that $Pr(W_{\beta} = 0)= Pr(W = 0)$ and for all $w>0$
$Pr(W_{\beta} \leq -w)=2Pr(W \geq w)/(1+\beta)$ and $Pr(W_{\beta} \geq w)=2\beta Pr(W \geq w)/(1+\beta)$, or equivalently for all $w > 0$
\begin{eqnarray}
\label{pop_symm}
Pr(W_{\beta} \geq w)=\beta Pr(W_{\beta} \leq -w).
\end{eqnarray}
Condition~(\ref{pop_symm}) is referred to as population weighted symmetry~\cite{wolfe1974characterization}. For a case of $\beta=1$,~(\ref{pop_symm}) reduces to a standard symmetric distribution and corresponds to the assumption~4 of~\citet{bernstein2018signsgd}. For
$\beta \neq 1$ (\ref{pop_symm}) describes a class of asymmetric distributions~\cite{rosenbaum2009amplification}. As such, ~(\ref{pop_symm}) allows us to consider a broader class of distributions than distributions which are symmetric around the mean as in the case of~\citet{bernstein2018signsgd}.}


{\bf Assumption~1} Gradient is Lipschitz continuous for each agent $k=1, \dots K$ and $L>0$ 
$$
||\nabla f_k(x)-\nabla f_k(y)|| \leq L||x-y||, \quad \forall x, y \in \mathbb{R}^d.
$$

{\bf Assumption~2} Variance for each agent $k=1, \dots, K$ is bounded,
$$
\mathbb{E}_{D_k} ||\nabla f_k(x, \xi^t_k)-\nabla f_k(x)|| \leq \sigma^2, \quad \forall x \in \mathbb{R}^d, \forall k \in \mathbb{Z}^+
$$

{\bf Assumption~3} Random variables $\xi^t_k$ are independent for all $k,t \in \mathbb{Z}^{+}$.


{\bf Theorem~1} (Convergence Rate)
  Let for all $i,k, t\in \mathbb{Z}^+$, $0\leq Pr(1- \IND |\digamma_t)\leq p_0 <0.25$, $0<\nu\leq (1-p_0)/L$ and $E||w^k_t||<M$, where $M>0$ is a universal clipping upper bound. 
Then under Assumptions 1-3, we have the following convergence rate for our robust learning rate scheme
$${\footnotesize
\frac{1}{T}\sum_{t=0}^{T-1}
\mathbb{E}||\nabla f(\hat{w}_t)||^2
\leq \frac{2}{\eta T} (f(\hat{w}_0)-f^{*})\nonumber + L^2M^2 +\frac{L\eta\sigma^2}{n},
}
$$
where $\hat{w}_t=1/n\sum_{k=1}^n w^k_t$.

See Appendix~\ref{app:proof} for the proof of the theorem.

\section{Experiments}\label{sec:exps}
In this section, we first illustrate the performance of our defense, and then provide some empirical justification for its effectiveness via experimental evaluation.
Our implementation is done using PyTorch~\cite{PyTorch}, and the code is available at \textit{https://github.com/TinfoilHat0/Defending-Against-Backdoors-with-Robust-Learning-Rate}.

\begin{figure}[!b]
  \begin{subfigure}[b]{0.5\textwidth}
  \center
    \includegraphics[scale=0.30]{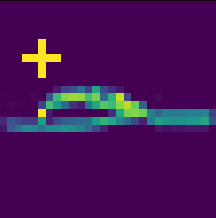}
     \includegraphics[scale=0.30]{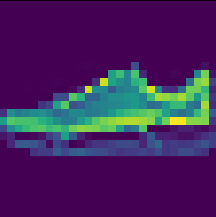}
     \caption{}
  \end{subfigure}
  \begin{subfigure}[b]{0.5\textwidth}
  \center
     \includegraphics[scale=0.30]{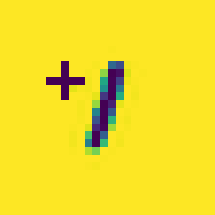}
     \includegraphics[scale=0.30]{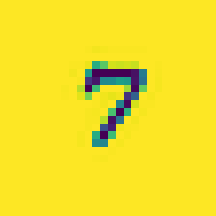}
     \caption{}
  \end{subfigure} 
  \caption{Samples from trojaned base classes and corresponding target classes. Trojan pattern is a 5-by-5 plus sign that is put to the top-left of objects. For \iid case (a), backdoor task is to make model classify trojaned sandals as sneakers. For \niid case (b), it is to make model classify trojaned digit 1s as digit 7s. Note that original images are in grayscale, these figures are normalized as they appear in training/validation dataset. We also repeat the experiments we present here under \emph{three additional trojan patterns} and report the results in Appendix.}\label{fig:trojaned_samples}
\end{figure}

The general setting of our experiments are as follows: we simulate FL for $R$ rounds among $K$ agents where $F$ fraction of them are corrupt.  The backdoor task is to make the model misclassify instances from a \emph{base class} as \emph{target class} by using trojan patterns. That is, a model having the backdoor classifies instances from base class with trojan pattern as target class (see Figure \ref{fig:trojaned_samples}). To do so, we assume an adversary who corrupts the local datasets of corrupt agents by adding a trojan pattern to $P$ fraction of base class instances and re-labeling them as target class. Other than that, adversary cannot view and modify updates of honest agents, or cannot influence the computation done by honest agents and the aggregation server. At each round, the server uniformly samples $C\cdot K$ agents for training where $C \leq 1$. These agents locally train for $E$ epochs with a batch size of $B$ before sending their updates. Upon receiving and aggregating updates, we measure three key performance metrics of the model on validation data: validation accuracy, base class accuracy and backdoor accuracy. Validation and base class accuracies are computed on the validation data that comes with the used datasets, and the backdoor accuracy is computed on a poisoned validation data that is constructed by (i) extracting all base class instances from the original validation data, and (ii) adding them the trojan pattern and re-labeling them as the target class. We measure the performance of the following five aggregation methods: (i) FedAvg (equation~\ref{eqn:fedavg}), (ii) FedAvg with \textit{our proposed robust learning rate scheme:} RLR (equation~\ref{eqn:fedavg_robust}), (iii) coordinate-wise median (comed), (iv) FoolsGold (equation~\ref{eqn:fgoold}), and (v) sign aggregation (equation~\ref{eqn:sign_agg}).
We also measure the performance of these aggregations under the proposed defense in
~\cite{sun2019really}, i.e., combining aggregations with weight-clipping and noise addition, to see if these techniques provide any robustness for each aggregation under our attack setting.
Furthermore, in Appendix, we provide results when comed and sign aggregation are combined with RLR. 

When there is a $L_2$ clipping threshold $M$ on updates, we assume $M$ is public and every agent runs projected gradient descent to minimize their losses under this restriction, \ie, an agent ensures his update's $L_2$ norm is bounded by $M$ by monitoring the $L_2$ norm of his model during training and clips its weights appropriately. Finally we use the same model as in~\cite{sun2019really}, a 5-layer convolutional neural network consisting of about 1.2M parameters with the following architecture: two layers of convolution, followed by a layer of max-pooling, followed by two fully-connected layers with dropout. Hyperparameters used in all experiments can be found in Appendix.

\begin{figure*}[!h]
\includegraphics[width=\textwidth]{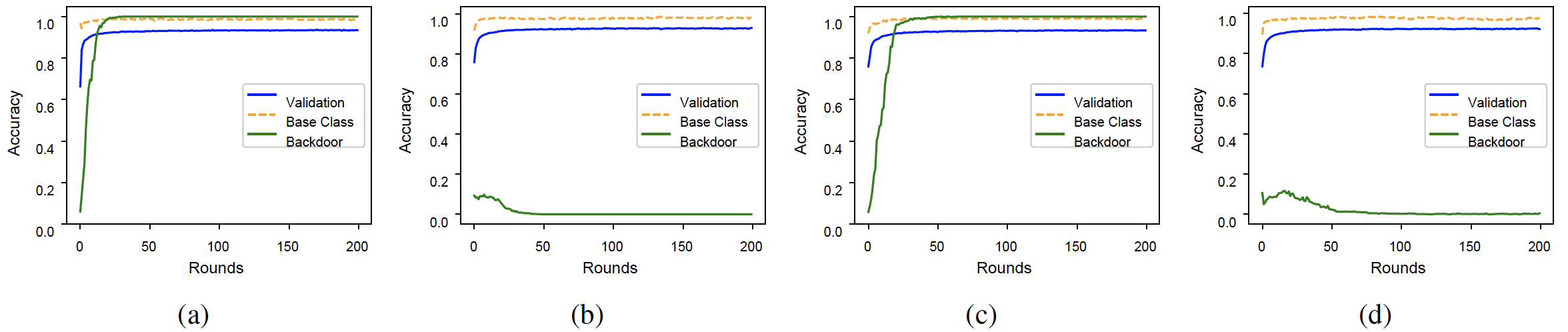}
  \caption{Training curves for FedAvg and FedAvg with RLR in \iid setting. From left-to-right: (a) FedAvg, (b) FedAvg with RLR, (c) FedAvg under clipping\&noise, (d) FedAvg with RLR under clipping\&noise. As can be seen, FedAvg is weak against the attack even with clipping\&noise. On the other hand, FedAvg with RLR prevents the backdoor with or without clipping\&noise. Using clipping and noise addition could be a desirable property in contexts where differential privacy is applied, or against attackers who try to make the model diverge by sending arbitrarily large values.}
  \label{fig:fedavg_iid}
\end{figure*}

\paragraph{IID Setting}
We start with a setting where data is distributed in \iid fashion among agents. Concretely, we use the Fashion MNIST~\cite{xiao2017fashionmnist} dataset, and give each agent an equal number of samples from the training data via uniform sampling.
In Figure~\ref{fig:fedavg_iid}, we plot the training curves of FedAvg, and FedAvg with RLR, and report the final accuracies reached in each setting in Table~\ref{table:results}. Results reported in Table~\ref{table:results} shows that, compared to baselines, our proposed RLR scheme provides significant protection against the backdoor attacks.

\begin{table*}[h!]
\centering

\begin{tabular}{ c c c c c c c } 
 \hline
 Aggregation & $M$ & $\sigma$ & Backdoor (\%) & Validation (\%) & Base (\%) \\ 
 \hline
 FedAvg-\emph{No Attack} & 0 & 0 & 1 & \textbf{93.5} & 98.5 \\
 FedAvg & 0 & 0 & 100 & 93.4 & 98.5 \\
 FedAvg & 4 & 1e-3 & 100 & 93.2 & \textbf{99.1} \\
 FoolsGold & 0 & 0 & 100 & 93.1 & 98.9 \\
 FoolsGold & 4 & 1e-3 & 100 & 93.3 & 98.5 \\
  Comed & 0 & 0 & 100 & 92.8 & 99.0 \\
  Comed & 4 & 1e-3 & 99.5 & 92.8 & 98.4 \\
 Sign & 0 & 0 & 100 & 92.9 & 98.7 \\
 Sign & 4 & 1e-3 & 99.7 & 93.1 & 98.6 \\
  FedAvg with RLR & 0 & 0 & \textbf{0} & 92.9  & 98.3 \\
  FedAvg with RLR & 4 & 1e-3 & 0.5 & 92.2 & 97.4 \\
  \hline
\end{tabular}

\begin{tabular}{ c c c c c c c } 
 \hline
 Aggregation & $M$ & $\sigma$ & Backdoor (\%) & Validation (\%) & Base (\%) \\ 
 \hline
 FedAvg*-\textit{No Attack} & 0 & 0 & 21.1 & \textbf{98.6} & \textbf{99.1} \\
 FedAvg & 0 & 0 & 99.3 & 98.5 & 99.0 \\
 FedAvg & 0.5 & 1e-3 & 99.2 & 98.0 & 98.7 \\
 FoolsGold & 0 & 0 & 98.5 & 98.9 & 99.5 \\
 FoolsGold & 0.5 & 1e-3 & 99.1 & 97.9 & 98.6 \\
  Comed & 0 & 0 & 82.3 & 96.3 & 98.4 \\
  Comed & 0.5 & 1e-3 & 95.2 & 95.5 & 98.1 \\
 Sign & 0 & 0 & 99.8 & 97.6 & 98.7 \\
 Sign & 0.5 & 1e-3 & 99.7 & 97.8 & 98.5 \\
  FedAvg with RLR & 0 & 0 & 3.4 & 94.8 & 97.6 \\
  FedAvg with RLR & 0.5 & 1e-3 & \textbf{0.4} & 93.2 & 97.7 \\
  \hline
\end{tabular}
\caption{Final backdoor, validation and base class accuracies for different aggregations in \iid (top) and \niid (bottom) settings. Lowest backdoor, highest validation and base class accuracies are highlighted in bold. FedAvg-No Attack corresponds to our baseline where we use FedAvg with no attackers.
See Appendix for additional experiments under different combinations of $M$ and $\sigma$, and our justification for the chosen values.}
\label{table:results} 
\end{table*}

\begin{figure*}[!h]
\includegraphics[width=\textwidth]{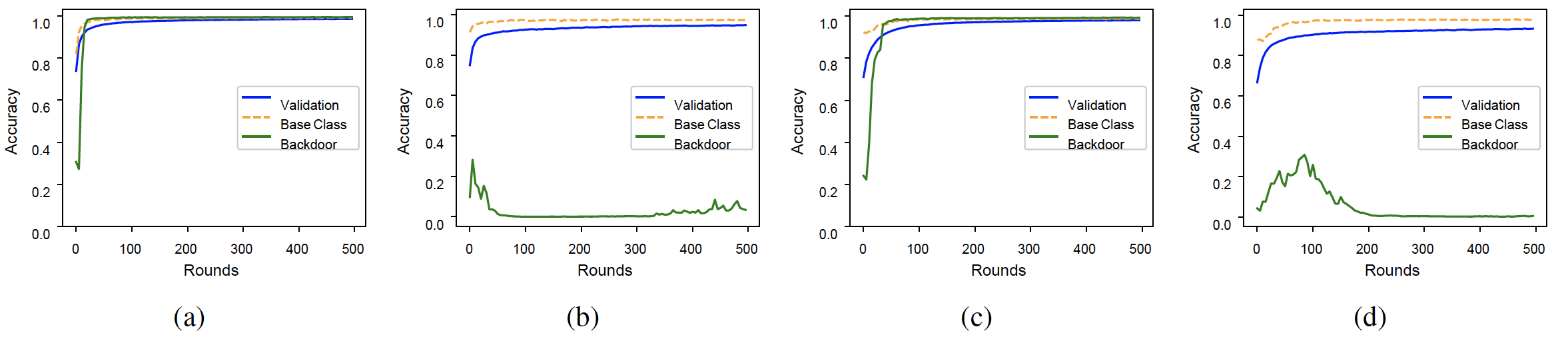}
  \caption{Plots for FedAvg and FedAvg with RLR in \niid setting. From left-to-right: (a) FedAvg, (b) FedAvg with RLR, (c) FedAvg under clipping\&noise, (d) FedAvg with RLR under clipping\&noise.}
  \label{fig:fedavg_niid}
\end{figure*}

\paragraph{Non-IID Setting}\label{niid_exps}
We now move on to a more realistic setting for FL in which data is distributed in \niid fashion among agents. For this setting, we use the Federated EMNIST dataset from the LEAF benchmark~\cite{caldas2018leaf}. In this dataset, digits 0-9 are distributed across 3383 users and each user has possibly a different distribution over digits. Similar to the \iid case, we plot training curves for FedAvg and FedAvg with RLR (Figure~\ref{fig:fedavg_niid}). Table~\ref{table:results} reports the final accuracy results for each setting. The results reported indicate that our defense provides the best protection with minimal degradation on the validation accuracy.

\paragraph{Removing Backdoor During Training}
During experiments, we observed that, FedAvg with RLR rate performs substantially better than other methods in terms of preventing the backdoor task, but it also reduces convergence speed. Therefore, we wonder if one can start without RLR, and then switch to RLR at some point during the training, e.g., when the model is about to converge, to clean any possible  backdoors from the model. Our experiments indicate that this is the case. In the interest of space, we provide results in Appendix, however they suggest that one can start without RLR and later switch to RLR when the model is about to converge, and/or a backdoor attack is suspected, to clean the model of backdoor during training. Overall, this improves the time to convergence when compared to using RLR right from the beginning.

\begin{figure*}[!t]
\includegraphics[width=\textwidth]{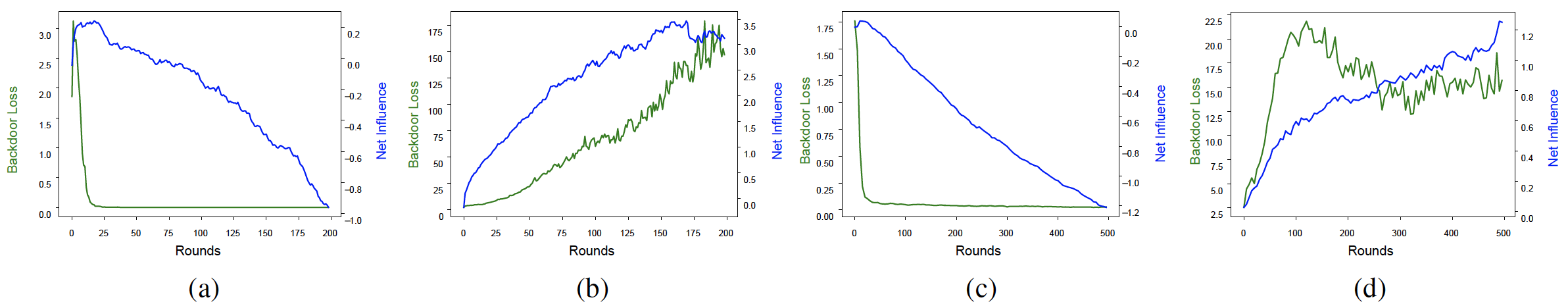}
  \caption{Results of parameter attribution experiments. From left-to-right: (a) FedAvg, (b) FedAvg with RLR in \iid setting, and (c) FedAvg, (d) FedAvg with RLR in \niid setting. Net influence is the cumulative sum of differences between the influences of honest agents and the adversarial agents for the mapping of trojaned samples. As can be seen, net influence is loosely correlated with the backdoor loss. With RLR, net influence is positive, indicating that honest agents' influence is greater than adversarial agents. This causes backdoor loss to increase, and hence, preventing the backdoor. On the other hand, without RLR, net influence quickly becomes negative and backdoor loss decreases. This results in a successful backdoor attack.}
  \label{fig:param_attr}
\end{figure*}

\paragraph{Analyzing Our Defense via Parameter and Feature Attributions}
We now aim to explain why our defense works and provide some empirical justification for its effectiveness. First, recall our conjecture from Section~\ref{sec:robustLR} where we basically argue that the adversary has to overcome the influence of honest agents to embed the backdoor to model. More concretely, in our scenario, adversary tries to map the base class instances with trojan pattern to the target class (adversarial mapping) where as honest agents try to map them to the base class (honest mapping). If we had a way to quantify the influence of agents on the model, regarding the mapping of trojaned inputs, we would expect the model to exhibit the backdoor if the influence of adversary is greater than of the total influence of honest agents. Given that, we designed a simple experiment to quantify the influence of agents, and test this conjecture empirically. In the interest of space,  we defer the details of this experiment to Appendix, but we note that it is mainly based on doing \emph{parameter attribution} on the model to find out which parameters are most important to adversarial/honest mapping, and then tracking how they are updated over the rounds. In Figure~\ref{fig:param_attr}, we can see that with RLR, honest agents' influence overcome the adversarial agents' for the backdoor task.

\begin{figure}[!t]
\center
  \begin{subfigure}{\columnwidth}
    \center
    \includegraphics[scale=0.2]{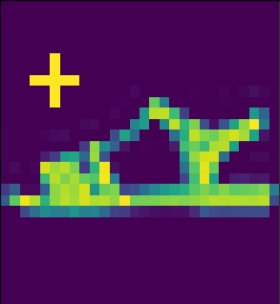}
    \includegraphics[scale=0.2]{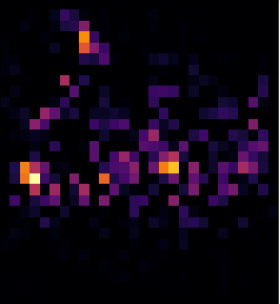}
    \includegraphics[scale=0.2]{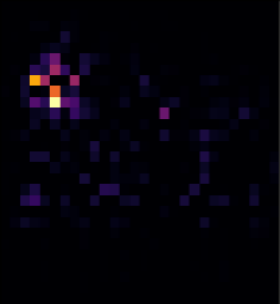}
    \includegraphics[scale=0.2]{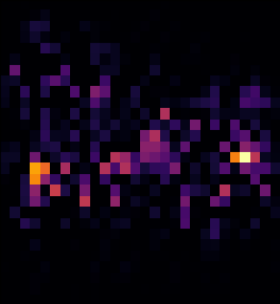}
    \caption{IID setting}
    \end{subfigure}
 
  \begin{subfigure}{\columnwidth}
    \center
    \includegraphics[scale=0.2]{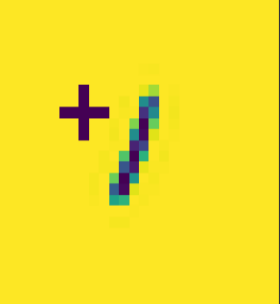}
    \includegraphics[scale=0.2]{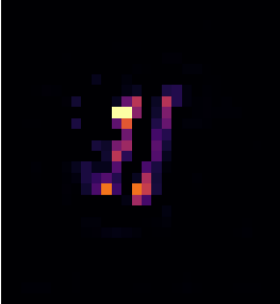}
    \includegraphics[scale=0.2]{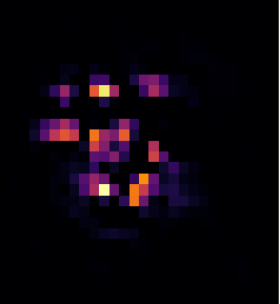}
    \includegraphics[scale=0.2]{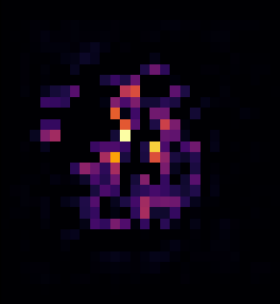}
    \caption{Non-IID setting}
    \end{subfigure}

 \includegraphics[scale=.5]{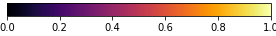}
 \caption{Feature maps (FM) for \iid and \niid settings on a trojaned sample given by Gradient SHAP~\cite{gradshap}. Leftmost image is the sample input from poisoned validation data, and to its right we present FMs in the following order: FM of model trained using FedAvg without any attack, FM of model trained using FedAvg under attack, FM of model trained using FedAvg with RLR under attack. For no attack case, important pixels are either  on or around the actual objects. For \iid setting, model predicts the sample correctly as sandals with $100\%$ confidence, and for \niid, model predicts the digit $1$ with $99.2\%$ confidence. For no defense scenario, we can see that model's attention has shifted towards the trojan pattern. This is especially very visible for \iid setting where the model almost completely focuses on the trojan. In \iid case, model predicts the sample as sneakers with $100\%$ confidence, and in \niid case, model predicts the digit as $7$ with $91.2\%$ confidence. Finally, we see that with robust learning rate, the model's attention has been shifted back to the actual objects to some extent. Now, model predicts the sample as sandals with $100\%$ confidence in \iid case, and it predicts the digit as $1$ with $91.2\%$ confidence in \niid case.}
  \label{fig:feature_maps}
  \end{figure}

Second, we do a \emph{feature attribution} experiment which is concerned with discovering features of an input that are important to a model's prediction.  Particularly, we pick an arbitrary sample from our poisoned validation set that is correctly classified (as base class) by the model when it is trained with FedAvg with RLR, but incorrectly classified (as target class) when it is trained FedAvg. Figure~\ref{fig:feature_maps} illustrates that, resulting feature maps on no attack and with our defense scenario are similar. This shows, our defense successfully prevents the model from focusing on the trojan pattern.

\paragraph{Distributed Backdoor Attacks}
Finally, we briefly test our defense against a recent, novel type of backdoor attack introduced in~\cite{xie2019dba}.
The main idea of this attack is to partition the pixels of a trojan between the agents of the adversary, and through that, ensuring the resulting malicious updates to be less different than honest' updates to make attack more stealthy. For example, if adversary has four agents, the plus pattern can be partitioned accross these four agents such that, each adversarial agent applies only a vertical/horizontal part of the plus. In case the backdoor is successful, the model would still misclassify the samples with the complete plus pattern.
We test this attack only against FedAvg with RLR, as other defenses already fail on default backdoor attacks, on CIFAR10 dataset~\cite{cifar10}. Table~\ref{tab:cifar10} indicates our defense performs well against distributed backdoor attacks too.

\setlength{\tabcolsep}{2pt}
\begin{table}[!ht]
\centering
\begin{tabular}{  c c c c  } 
 \hline
 Aggregation & Backdoor (\%) & Validation (\%) & Base (\%) \\ 
\hline
 FedAvg-\textit{No Attack} & 6.6 & 79.0 & 89.4 \\
 FedAvg & 88.6 & 79.4 & 87.5 \\
 FedAvg with RLR & 9.0 & 77.5 & 87.8 \\
  \hline
 \end{tabular}

\begin{tabular}{  c c c c  } 
 \hline
 Aggregation & Backdoor (\%) & Validation (\%) & Base (\%) \\ 
 \hline
 FedAvg-\emph{No Attack} &  6.3 & 76.6 & 87.7 \\
 FedAvg &  61.7 & 76.6 & 78.2 \\
 FedAvg with RLR &  8.5 & 71.8 & 83.3 \\
 \hline
 \end{tabular}

\caption{Backdoor attack on \iid-partitioned CIFAR10. Backdoor task is to classify dogs (base class) with plus pattern as horses (target class). Top table is for regular backdoor attack, and bottom table is for distributed backdoor attack where plus pattern is partitioned to 4 adversarial agents out of 40 agents. See Appendix for details.}
\label{tab:cifar10}
\end{table}
\section{Discussion}\label{sec:discussion}
Our experiments show that our approach significantly reduces the effectiveness of trojan pattern backdoor attacks. One can wonder that, how it performs with respect to the so-called semantic backdoors (a.k.a label-flipping) attacks. In these attacks, the adversary simply flips the label of the base class instances to a desired target label without adding a trojan pattern. In FL setting, it has been shown that successfully carrying such attacks require \emph{boosting}~\cite{arxiv:2018:fedlens}. That is, after training on a poisoned dataset, adversary has to multiply the resulting update with a large constant to overcome the effect of honest agents. Naturally, this results in adversarial updates having a large norm, and as shown in~\cite{sun2019really}, weight-clipping and noise addition significantly deters these attacks. Since our defense is compatible with clipping and noise addition, it can also deter such attacks. In fact, our experiment show that, trojan backdoors are strictly more powerful than semantic backdoors in FL context as an adversary does not need to use boosting with them.

Finally, we ask if an adversary can devise a clever attack. At a high level, as long as the $\theta$ parameter is set appropriately, and adversary's local loss function differs from the honest against, the scheme will try to move the model from the directions the adversarial update specifies. Adversary could try to make his loss function more in-line with honest agents' via some modification, but then this will likely result in his attack losing effectiveness. We emphasize that our approach does not \say{magically} finds the adversary, and negates his update by multiplying it with $-\eta$, so the adversary cannot by-pass our defense just by negating his loss.

\section{Conclusion} \label{sec:conc}
In this work, we studied FL from an adversarial perspective, and constructed a simple defense mechanism, particularly against backdoor attacks. The key idea behind our defense was adjusting the aggregation server's learning rate, per dimension and per round, based on the sign information of agents' updates. Through experiments we present above and in Appendix, we illustrate that our defense reduces backdoor accuracy substantially with a minimal degradation in the overall validation accuracy. Overall, it outperforms some of the recently proposed defenses in the literature. As a final comment, we believe the insights behind our defense are also related to training in \niid setting, even in the presence of no adversaries. Because, the differences in local distributions can cause updates coming from different agents to steer the model towards different directions over the loss surface. As a future work, we plan to analyze how RLR influences performance of models trained in different \niid settings.

\section*{Acknowledgments}
The research reported herein was supported in part by NIH award 
1R01HG006844, NSF awards, CNS-1633331, CNS-1837627, OAC-1828467, 
IIS-1939728, DMS-1925346, CNS-2029661 and ARO award W911NF-17-1-0356.

\bibliography{main}
\appendix
\appendixpage
\section{Hyperparameters of Experiments}\label{app:hyperparams}
We remind the notation we introduced at the beginning of Section~\ref{sec:exps} and report the hyperparameters of our experiments. We also briefly discuss our choices. 

\begin{itemize}
    \item{R: Number of rounds }
    \item{K: Total number of agents}
    \item{F: Fraction of corrupt agents}
    \item{P: Fraction of trojaned samples in a corrupt agent's dataset}
    \item{C: Fraction of selected agents for training in a round}
    \item{E: Number of epochs in local training} 
    \item{B: Batch size of local training}
    \item{$\eta$: Server's learning rate} 
    \item{$\theta$: Threshold for RLR (see equation~\ref{eqn:rlrTheta})}
\end{itemize}

\begin{table}[h!]
\centering
\label{tab:hyperparamsIID}
\resizebox{0.5\columnwidth}{!}{%
\begin{tabular}{  c c c c c c c c } 
 \toprule
 R & K & F & P & C & E & B \\
 \midrule
200 & 10 & 0.1 & 0.5 & 1 & 2 & 256 \\
 \bottomrule
 \end{tabular}
 }
 \caption{Hyperparameters for all \iid experiments. In addition to what is presented in table, we set $\eta$ to 1e-3 when sign aggregation is used and to 1 otherwise. Finally, $\theta$ is set to 4 when RLR used.}
 \end{table}

\begin{table}[h!]
\centering
\label{tab:hyperparamsIID}
\resizebox{0.5\columnwidth}{!}{%
\begin{tabular}{  c c c c c c c c } 
 \toprule
 R & K & F & P & C & E & B \\
 \midrule
500 & 3383 & 0.1 & 0.5 & 0.01 & 10 & 64 \\
 \bottomrule
 \end{tabular}}
 \caption{Hyperparameters for all \niid experiments. In addition to what is presented in table, we set $\eta$ to 1e-3 when sign aggregation is used and to 1 otherwise. Finally, $\theta$ is set to 7 when RLR used.}
 \end{table}
In both cases, we set E and B to some values that loosely help us to run as many experiments as quickly as possible in our system. F was arbitrarily fixed to 0.1 so as the values for C. We set P to 0.5 after trying different values and observing that the backdoor accuracy rises the quickest under that value to simulate a strong adversary. Setting the value of $\theta$ is non-trivial. Technically, it could be any value between $K.F+1$,
$K - K.F$. In our experiments, setting it to $4$ in \iid setting seemed to provide us the best trade-off between backdoor prevention and the drop in validation accuracy. For \niid setting, in expectation, we had $3$ corrupt agents per round, and setting $\theta$ to $7$ gave us a similar trade-off as in \iid case.

Finally, hyperparameters for distributed backdoor attack experiment on CIFAR10 (see Figure~\ref{tab:cifar10}) is given in Table~\ref{tab:hyperparamsCifar10}.

\begin{table}[h!]
\centering
\resizebox{0.5\columnwidth}{!}{%
\begin{tabular}{  c c c c c c c c } 
 \toprule
 R & K & F & P & C & E & B \\
 \midrule
100 & 40 & 0.1 & 0.5 & 1 & 2 & 256 \\
 \bottomrule
 \end{tabular}}
 \caption{Hyperparameters for all CIFAR10 experiments. In addition to what is presented in table, we set $\eta$ to 1 Finally, $\theta$ is set to 8 when RLR used. Under distributed backdoor attack, four lines of the plus pattern is partitioned across four adversarial agents.}
 \label{tab:hyperparamsCifar10}
 \vspace{-15pt}
 \end{table}

\section{Parameter Attribution Experiment}\label{app:influence_exp}
We quantified influences of corrupt/honest agents as follows. After each round, we find the 100 most important parameters for adversarial, and honest mapping by computing the empirical Fisher Information Matrix (FIM) as done in~\cite{shoham2019overcoming}. Particularly, we compute the diagonal of FIM on the trojaned samples, labeled as target class, and take top 100 values for adversarial mapping. We do the same by computing FIM on trojaned samples labeled as base class to find out the most influential parameters for honest mapping. Then, due to RLR, some of these top 100 parameters are updated in a way to minimize the loss, and some of them are updated in a way to maximize the loss. Let $S_1, S_2$ be those which are updated to minimize the loss for adversarial and honest mapping, respectively. Also let $S_3, S_4$ be those which are updated to maximize the loss for adversarial and honest mapping, respectively. Then, we quantified the net adversarial influence as $I_{adv} = \Vert S_1 \setminus S_2\Vert_2$ - $\Vert S_3 \setminus S_4 \Vert_2$ and the net honest influence as $I_{hon}= \Vert S_2 \setminus S_1\Vert_2$ - $\Vert S_4 \setminus S_3 \Vert_2$. Finally, the net influence is then given by $I_{hon} - I_{adv}$ which is plotted in blue in Figure~\ref{fig:param_attr}.

\section{Extra Experiments}\label{app:extra_exps}

\subsection{Removing Backdoor During Training}
\label{sec:cleanse}
\begin{figure*} 
\centering
  \begin{subfigure}[b]{0.3\textwidth}
    \includegraphics[width=\textwidth]{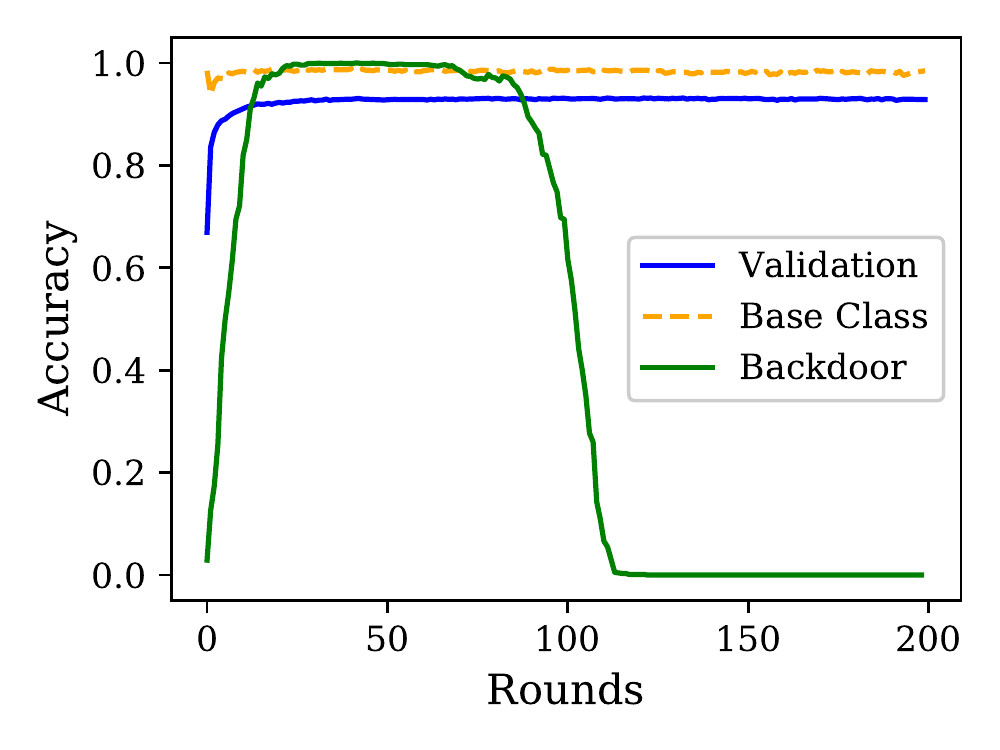}
    \caption{IID setting}
    \label{fig:f5}
  \end{subfigure}
  \hfill
  \begin{subfigure}[b]{0.3\textwidth}
    \includegraphics[width=\textwidth]{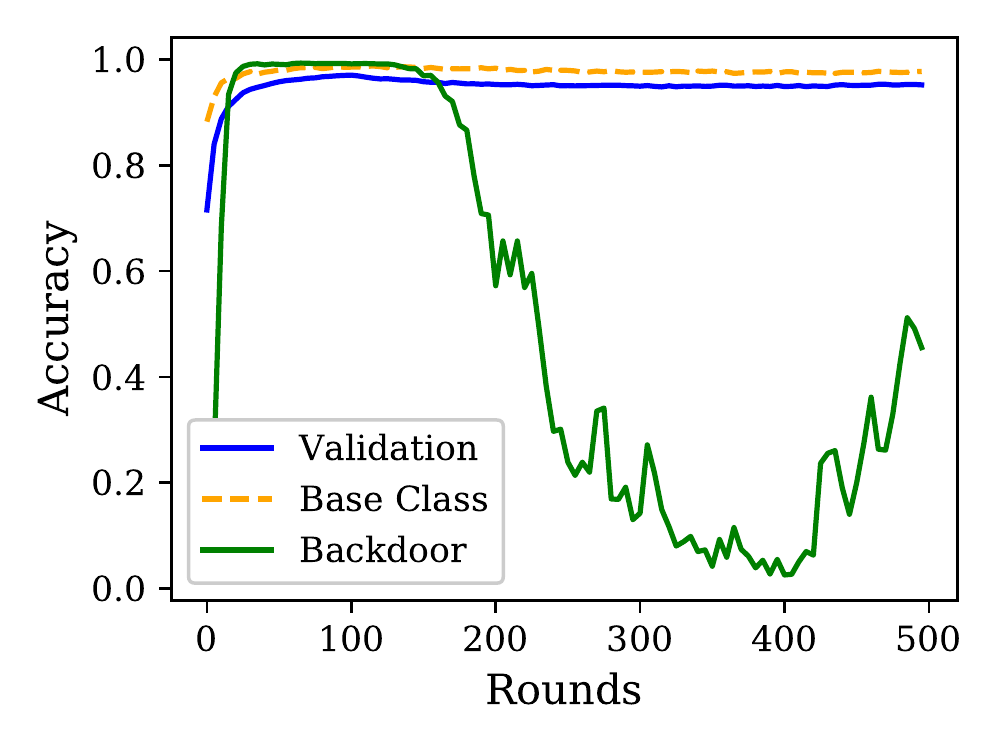}
    \caption{Non-IID setting, $\theta=7$}
    \label{fig:cleanse_niid_7}
  \end{subfigure}
  \hfill
  \begin{subfigure}[b]{0.3\textwidth}
    \includegraphics[width=\textwidth]{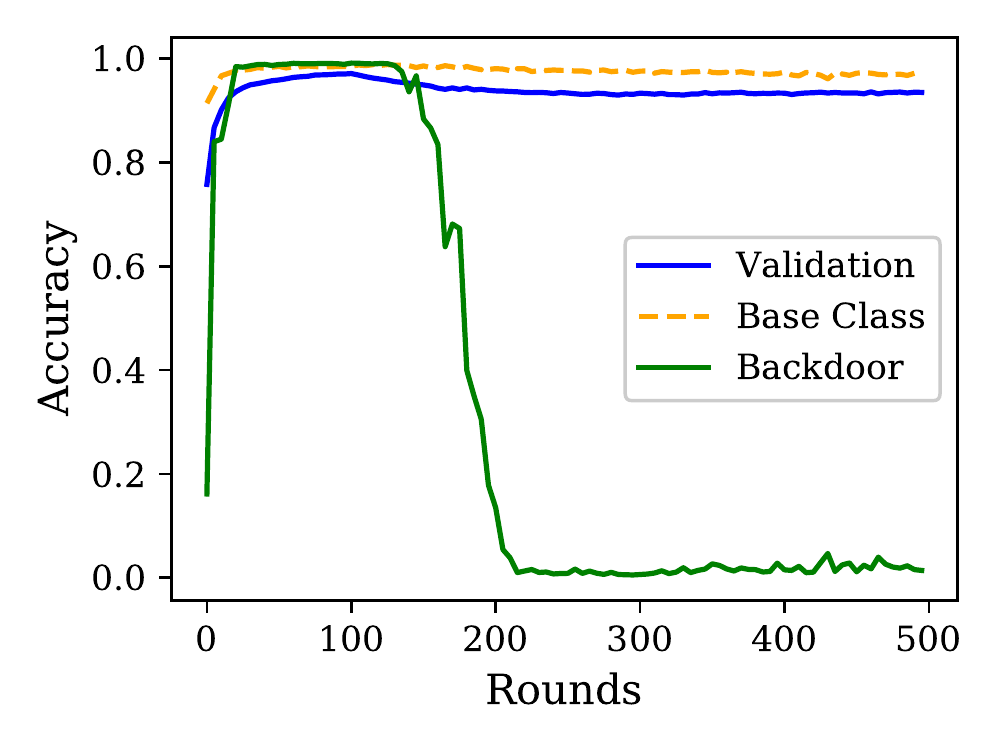}
    \caption{Non-IID setting, $\theta=8$}
    \label{fig:f6}
  \end{subfigure} 
  \caption{Cleaning the backdoor during training by activating the RLR when the model is about to converge. For \iid setting, we activate the RLR when the model's validation accuracy is above $93\%$. This occurs at round $41$, and at that round, backdoor accuracy is at $100\%$. At round 124, the backdoor accuracy is $0\%$ with a validation accuracy of $93.1\%$. Final validation and base class accuracies are $92.8\%$ and $98.5\%$ respectively. On the other hand, for \niid setting, we activate RLR when the validation accuracy is above $97\%$. However in (b), we observe that using the same RLR threshold (7) that we used in Section \ref{niid_exps} fails to prevent the backdoor now. Backdoor accuracy falls to $2\%$ from $99\%$ and then rises up to $45\%$ by round 500. Yet, if we increase the threshold to 8, we observe it performs substantially better in (c). In this case, final backdoor accuracy is $1.4\%$, and final validation and base class accuracies are $93.4\%$ and $97\%$, respectively.}
  \label{fig:cleansing}
  \vspace{-10pt}
\end{figure*}

See the comments on Figure~\ref{fig:cleansing}.

\subsection{Higher Corruption Percentages}
See the comments on Table~\ref{tab:higherCorrupt}.

\begin{table}[t!]
\centering
\resizebox{\columnwidth}{!}{%
\begin{tabular}{  c c c c c c } 
 \toprule
 Corruption (\%) & Aggregation & Backdoor (\%) & Validation (\%) & Base (\%) \\ 
 \midrule
 
 20 & FedAvg & 99.6 & 92.9 & 98.1 \\
 20 & FedAvg with RLR & 0 & 89.6 & 96.6 \\
 30 & FedAvg & 99.9 & 92.9 & 97.4 \\
 30 & FedAvg with RLR & 0 & 86.6 & 95.2 \\
 40 & FedAvg & 99.8 & 92.7 & 97.7 \\
 40 & FedAvg with RLR & 0 & 82.5 & 94.2 \\
 \bottomrule
 \end{tabular}
 }
 \resizebox{\columnwidth}{!}{%
\begin{tabular}{  c c c c c c } 
 \toprule
 Corruption (\%) & Aggregation & Backdoor (\%) & Validation (\%) & Base (\%) \\ 
 \midrule
 
 20 & FedAvg & 99.5 & 97.9 & 99.0 \\
 20 & FedAvg with RLR & 4.3 & 90.7 & 97.5 \\
 30 & FedAvg & 99.5 & 97.9 & 98.8 \\
 30 & FedAvg with RLR & 0.8 & 87.9 & 97.4 \\
 40 & FedAvg & 99.5 & 97.8 & 98.7 \\
 40 & FedAvg with RLR & 0.1 & 77.5 & 96.5 \\
 \bottomrule
 \end{tabular}
 }
 \caption{Performance of RLR with FedAvg for higher corruption rates. Top is the \iid setting, bottom is the \niid setting. For these experiments, we used 40 agents in \iid case. Furthermore, activating the RLR towards the end of the training as in~\ref{sec:cleanse} performed much better in \niid case as corruption rate increased. So, for \niid case, we activated RLR 50 epochs before the end of the training in these experiments. For \iid case, $\theta$ values were set to $12, 16, 20$, and for \niid case, they were $17, 22, 27$ for $20\%, 30\%, 40\%$ corruption, respectively. Note that, we might have been more conservative than needed when setting $\theta$ values for higher percentages. This could explain why backdoor accuracy gets lower as corruption increases in \niid case.}
 \label{tab:higherCorrupt}
 \end{table}

\subsection{Negating Loss Function Attack}
\label{app:negateAttack}
We briefly show that an adversary cannot by-pass our defense simply by negating the sign of his loss function. We explained, at an intuitive level, why such an attack would fail in the second paragraph of Section~\ref{sec:discussion}, and in Table~\ref{tab:negate_attack}, we provide the experimental confirmation.

\begin{table}[!t]
\captionsetup{font=small}
\centering
\resizebox{0.75\columnwidth}{!}{%
\begin{tabular}{ c c c c c c c } 
 \hline
 Setting &  Backdoor (\%) & Validation (\%) & Base (\%) \\ 
 \hline
 IID & 0.7 & 90.8 & 98.4 \\
 Non-IID & 3 & 92.2 & 98.0 \\
 \hline
\end{tabular}}
\caption{Results of what happens when adversary tries to by-pass RLR by negating his loss function. The aggregation is FedAvg with RLR. We can see this \say{attack} does not give anything useful to adversary for backdoor task when results are compared with Table~\ref{table:results}. However, since adversary tries to maximize his original loss function due to negation, this causes norms of his updates to be very large. So, we had to use clipping at the server-side with values of $M=4$ for \iid setting, and $M=0.5$ for \niid setting to prevent model from diverging.}
\label{tab:negate_attack}
\end{table}

\subsection{Experiments for all $M, \sigma$ combinations and Trojan Patterns}
\label{app:extraM}
\begin{figure*}
\centering
  \begin{subfigure}{\columnwidth}
  \center
    \includegraphics[scale=0.3]{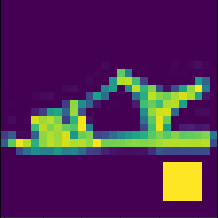}
    \includegraphics[scale=0.3]{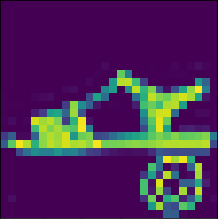}
    \includegraphics[scale=0.3]{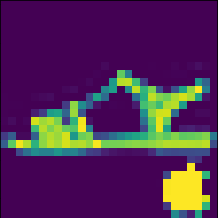}
   \caption{IID setting}
  \end{subfigure}
  \vfill
  \begin{subfigure}{\columnwidth}
  \center
    \includegraphics[scale=0.3]{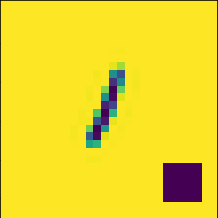}
    \includegraphics[scale=0.3]{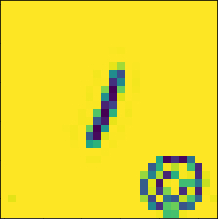}
    \includegraphics[scale=0.3]{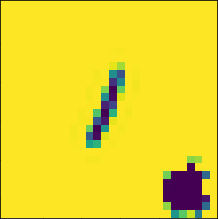}
    \caption{Non-IID setting}
    \end{subfigure}
  \caption{Extra trojans patterns from~\cite{liu2017trojaning} as applied to datasets we use.}
  \label{fig:extraTrojans}
\end{figure*}

Regarding our choice of $M,\sigma$ in \iid case, we observed $L_2$ norm of updates of honest agents during training in baseline which happened to be floating around 6. So, we ran experiments for $M=6, 4, 2$. For $\sigma$, we tried $\sigma=1e-4, 1e-3, 5e-3$ and stopped increasing it after observing that training becomes imbalanced at $5e-3$. 
For \niid case, $L_2$ norm of updates of honest agents were about $1.5-2$, so we have chosen $M=1, 0.5, 0.25$ and used the same $\sigma$ values from values from \iid case. In addition to the plus trojan pattern we used in the main body, we repeated our experiments under three more trojans which are the same as in~\cite{liu2017trojaning}: a square, a copyright logo, an Apple logo, placed to the bottom-right of objects (see Figure~\ref{fig:extraTrojans}). We provide explicit results for all possible settings for FedAvg in Tables~\ref{tab:appendixFedAvgIID} and ~\ref{tab:appendixFedAvgNIID}. For other aggregations, we provide the configuration for the setting where the backdoor accuracy is the lowest, for each trojan, for sake of brevity in Tables~\ref{tab:appendixComedSignIID} and~\ref{tab:appendixComedSignNIID}. Results indicate FedAvg combined with RLR outperforms the other techniques. However, comed and sign also perform well especially under square, copyright and Apple logo trojans in \niid setting (Table~\ref{tab:appendixComedSignNIID}). Another point to note is, sign aggregation seems to be performing well with RLR while comed sometimes performs better without it. 

\begin{table}[!t]
\centering

\resizebox{\columnwidth}{!}{%
\begin{tabular}{  c c c c c c c c } 
 \toprule
 Trojan Pattern & $M$ & $\sigma$ & RLR used? & Backdoor (\%) & Validation (\%) & Base (\%) \\ 
 \midrule
 Plus & 6 & 5e-3 & No & 10.5 & 85.8 & 94.0 \\
 Square & 4 & 5e-3 & No & 1.9 & 88.7 & 95.9 \\
 Copyright & 6 & 5e-3 & No & 5.7 & 85.5 & 95.5 \\
 Apple & 4 & 5e-3 & No & 4.4 & 88.6 & 95.9 \\
  \bottomrule
 \end{tabular}
 }
 \resizebox{\columnwidth}{!}{%
\begin{tabular}{  c c c c c c c c } 
 \toprule
 Trojan Pattern & $M$ & $\sigma$ & RLR used? & Backdoor (\%) & Validation (\%) & Base (\%) \\ 
 \midrule
 Plus & 6 & 5e-3 & Yes & 92.1 & 92.4 & 98.1 \\
 Square & 0 & 1e-3 & Yes & 4.9 & 98.2 & 92.6 \\
 Copyright & 6 & 5e-3 & Yes & 47.7 & 92.5 & 98.1 \\
 Apple & 0 & 1e-4 & Yes & 8.5 & 92.6 & 98.9 \\
  \bottomrule
 \end{tabular}}
  \caption{Results for comed/sign (top/bottom) in \iid setting under different trojans. }
 \label{tab:appendixComedSignIID}

 \vspace{-10pt}
 \end{table}
 
\begin{table}[t]

\resizebox{\columnwidth}{!}{%
\begin{tabular}{  c c c c c c c c } 
 \toprule
 Trojan Pattern & $M$ & $\sigma$ & RLR used? & Backdoor (\%) & Validation (\%) & Base (\%) \\ 
 \midrule
 Plus & 0.5 & 5e-3 & Yes & 10.5 & 94.8 & 98.1 \\
 Square & 0.5 & 1e-4 & Yes & 0.1 & 94.9 & 98.1 \\
 Copyright & 1 & 0 & No & 0.1 & 96.5 & 98.3 \\
 Apple & 0 & 5e-3 & Yes & 0.1 & 96.2 & 98.4 \\
  \bottomrule
 \end{tabular}}
 \qquad
 \resizebox{\columnwidth}{!}{%
\begin{tabular}{  c c c c c c c c } 
 \toprule
 Trojan Pattern & $M$ & $\sigma$ & RLR used? & Backdoor (\%) & Validation (\%) & Base (\%) \\ 
 \midrule
 Plus & 0.25 & 1e-3 & Yes & 54.2 & 94.6 & 98.3 \\
 Square & 0 & 0 & Yes & 0 & 95.4 & 98.3 \\
 Copyright & 0 & 1e-4 & Yes & 0 & 94.6 & 98.6 \\
 Apple & 0 & 0 & Yes & 0 & 95.4 & 98.5 \\
  \bottomrule
 \end{tabular}}
  \caption{Results for comed/sign (top/bottom) in \niid setting under different trojans.}
 \label{tab:appendixComedSignNIID}

 \end{table}

\begin{table*}
\caption{Results for FedAvg in \iid setting under different trojans. Top-left/right: plus/square, bottom-left/right: copyright/Apple logo.}
\resizebox{\columnwidth}{!}{%
\begin{tabular}{  c c c c c c c c } 
 \toprule
 Aggregation & $M$ & $\sigma$ & RLR used? & Backdoor (\%) & Validation (\%) & Base (\%) \\
 \midrule
 FedAvg* & 0 & 0 & No & 1 & 93.5 & 98.5 \\
 FedAvg & 0 & 0 & No & 100 & 93.4 & 98.5 \\
 FedAvg & 0 & 0 & Yes & 0 & 92.9 & 98.3 \\
  
 FedAvg & 0 & 1e-4 & No & 100 & 93.3 & 98.8 \\
 FedAvg & 0 & 1e-4 & Yes & 0 & 92.5 & 98.0 \\
 
 FedAvg & 0 & 1e-3 & No & 100 & 93.2 & 98.4 \\
 FedAvg & 0 & 1e-3 & Yes & 0 & 92.4 & 98.0 \\
  
 FedAvg & 0 & 5e-3 & No & 100 & 93.2 & 98.6 \\
 FedAvg & 0 & 5e-3 & Yes & 0 & 92.5 & 98.4 \\
 
 FedAvg & 2 & 0 & No & 100 & 93.3 & 98.8 \\
 FedAvg & 2 & 0 & Yes & 0 & 92.7 & 98.0 \\
 
 FedAvg & 2 & 1e-4 & No & 100 & 93.5 & 99.2 \\
 FedAvg & 2 & 1e-4 & Yes & 0 & 92.9 & 98.8 \\
  
 FedAvg & 2 & 1e-3 & No & 100 & 93.4 & 98.8 \\
 FedAvg & 2 & 1e-3 & Yes & 0 & 92.6 & 98.3 \\
 
 FedAvg & 2 & 5e-3 & No & 99.8 & 91.2 & 97.6 \\
 FedAvg & 2 & 5e-3 & Yes & 4 & 89.6 & 95.5 \\
  
 FedAvg & 4 & 0 & No & 100 & 93.4 & 99.0 \\
 FedAvg & 4 & 0 & Yes & 0 & 93.3 & 98.3 \\
  
 FedAvg & 4 & 1e-4 & No & 100 & 93.6 & 99.0 \\
 FedAvg & 4 & 1e-4 & Yes & 0 & 92.9 & 98.2 \\
 
 FedAvg & 4 & 1e-3 & No & 100 & 93.2 & 99.1 \\
 FedAvg & 4 & 1e-3 & Yes & 0.5 & 92.2 & 97.4 \\
  
 FedAvg & 4 & 5e-3 & No & 99.0 & 89.0 & 96.3 \\
 FedAvg & 4 & 5e-3 & Yes & 3.1 & 87.2 & 94.0 \\
 
 FedAvg & 6 & 0 & No & 100 & 93.4 & 98.5 \\
 FedAvg & 6 & 0 & Yes & 0 & 93.0 & 97.9 \\
  
 FedAvg & 6 & 1e-4 & No & 100 & 93.5 & 99.1 \\
 FedAvg & 6 & 1e-4 & Yes & 0 & 92.9 & 98.1 \\
 
 FedAvg & 6 & 1e-3 & No & 100 & 93.1 & 98.6 \\
 FedAvg & 6 & 1e-3 & Yes & 0.5 & 91.8 & 97.2 \\
  
 FedAvg & 6 & 5e-3 & No & 92.3 & 85.7 & 93.9 \\
 FedAvg & 6 & 5e-3 & Yes & 6.1 & 83.9 & 93.1 \\
 \bottomrule
 \end{tabular}}
  \vspace{10 mm}
 \qquad
\resizebox{\columnwidth}{!}{%
\begin{tabular}{  c c c c c c c c } 
 \toprule
 Aggregation & $M$ & $\sigma$ & RLR used? & Backdoor (\%) & Validation (\%) & Base (\%) \\
 \midrule
 FedAvg* & 0 & 0 & No & 0.7 & 93.2 & 99.0 \\
 FedAvg & 0 & 0 & No & 95.0 & 93.3 & 98.5 \\
 FedAvg & 0 & 0 & Yes & 0 & 92.6 & 98.5 \\
  
 FedAvg & 0 & 1e-4 & No & 94.9 & 93.4 & 98.3 \\
 FedAvg & 0 & 1e-4 & Yes & 0 & 92.4 & 98.3 \\
 
 FedAvg & 0 & 1e-3 & No & 95.9 & 93.4 & 98.8 \\
 FedAvg & 0 & 1e-3 & Yes & 0 & 92.5 & 97.9 \\
  
 FedAvg & 0 & 5e-3 & No & 94.0 & 93.6 & 99.0 \\
 FedAvg & 0 & 5e-3 & Yes & 0 & 92.6 & 98.3 \\
 
 FedAvg & 2 & 0 & No & 96.5 & 93.4 & 98.9 \\
 FedAvg & 2 & 0 & Yes & 0 & 92.7 & 98.4 \\
 
 FedAvg & 2 & 1e-4 & No & 94.2 & 93.4 & 98.7 \\
 FedAvg & 2 & 1e-4 & Yes & 0 & 92.8 & 98.1 \\
  
 FedAvg & 2 & 1e-3 & No & 93.8 & 93.6 & 99.3 \\
 FedAvg & 2 & 1e-3 & Yes & 0 & 92.6 & 98.0 \\
 
 FedAvg & 2 & 5e-3 & No & 42.6 & 91.2 & 97.5 \\
 FedAvg & 2 & 5e-3 & Yes & 1.4 & 89.8 & 96.3 \\
  
 FedAvg & 4 & 0 & No & 95.0 & 93.5 & 98.8 \\
 FedAvg & 4 & 0 & Yes & 0.1 & 93.2 & 98.3 \\
  
 FedAvg & 4 & 1e-4 & No & 93.7 & 93.5 & 99.0 \\
 FedAvg & 4 & 1e-4 & Yes & 0 & 93.4 & 97.7 \\
 
 FedAvg & 4 & 1e-3 & No & 94 & 93.4 & 99.0 \\
 FedAvg & 4 & 1e-3 & Yes & 0.5 & 92.3 & 97.5 \\
  
 FedAvg & 4 & 5e-3 & No & 33.6 & 88.8 & 97.1 \\
 FedAvg & 4 & 5e-3 & Yes & 3.1 & 87.2 & 94.0 \\
 
 FedAvg & 6 & 0 & No & 93.2 & 93.4 & 98.5 \\
 FedAvg & 6 & 0 & Yes & 0 & 93.1 & 97.7 \\
  
 FedAvg & 6 & 1e-4 & No & 94 & 93.4 & 99.2 \\
 FedAvg & 6 & 1e-4 & Yes & 0 & 93.1 & 98.5 \\
 
 FedAvg & 6 & 1e-3 & No & 94.3 & 92.9 & 98.5 \\
 FedAvg & 6 & 1e-3 & Yes & 0.5 & 91.4 & 98.1 \\
  
 FedAvg & 6 & 5e-3 & No & 25.4 & 85.3 & 94.5 \\
 FedAvg & 6 & 5e-3 & Yes & 6.1 & 83.3 & 88.2 \\
 \bottomrule
 \end{tabular}}
 
\resizebox{\columnwidth}{!}{%
\begin{tabular}{  c c c c c c c c } 
 \toprule
 Aggregation & $M$ & $\sigma$ & RLR used? & Backdoor (\%) & Validation (\%) & Base (\%) \\ 
 \midrule
 FedAvg* & 0 & 0 & No & 0.4 & 93.6  & 98.8 \\
 FedAvg & 0 & 0 & No & 98 & 93.3 & 98.8 \\
 FedAvg & 0 & 0 & Yes & 98.7 & 92.7 & 98.3 \\
  
 FedAvg & 0 & 1e-4 & No & 98.0 & 93.3 & 98.8 \\
 FedAvg & 0 & 1e-4 & Yes & 0 & 92.9 & 98.2 \\
 
 FedAvg & 0 & 1e-3 & No & 97.6 & 93.6 & 98.9 \\
 FedAvg & 0 & 1e-3 & Yes & 0 & 92.6 & 97.9 \\
  
 FedAvg & 0 & 5e-3 & No & 98.9 & 93.0 & 99.0 \\
 FedAvg & 0 & 5e-3 & Yes & 0 & 93.0 & 97.6 \\
 
 FedAvg & 2 & 0 & No & 98.1 & 93.4 & 98.9 \\
 FedAvg & 2 & 0 & Yes & 0 & 92.7 & 97.8 \\
 
 FedAvg & 2 & 1e-4 & No & 98.2 & 93.7 & 98.7 \\
 FedAvg & 2 & 1e-4 & Yes & 0 & 92.7 & 98.4 \\
  
 FedAvg & 2 & 1e-3 & No & 97.4 & 93.4 & 98.8 \\
 FedAvg & 2 & 1e-3 & Yes & 0 & 92.6 & 97.5 \\
 
 FedAvg & 2 & 5e-3 & No & 93.2 & 91.2 & 97.1 \\
 FedAvg & 2 & 5e-3 & Yes & 1.5 & 90.0 & 96.6 \\
  
 FedAvg & 4 & 0 & No & 98.7 & 93.6 & 99.2 \\
 FedAvg & 4 & 0 & Yes & 0.1 & 93.1 & 98.1 \\
  
 FedAvg & 4 & 1e-4 & No & 97.5 & 93.4 & 99.0 \\
 FedAvg & 4 & 1e-4 & Yes & 0 & 93.1 & 97.9 \\
 
 FedAvg & 4 & 1e-3 & No & 98.0 & 93.0 & 98.8 \\
 FedAvg & 4 & 1e-3 & Yes & 0 & 92.1 & 97.9 \\
  
 FedAvg & 4 & 5e-3 & No & 94.7 & 88.7 & 95.3 \\
 FedAvg & 4 & 5e-3 & Yes & 7.6 & 86.8 & 90.5 \\
 
 FedAvg & 6 & 0 & No & 98.7 & 93.4 & 98.5 \\
 FedAvg & 6 & 0 & Yes & 0 & 93.1 & 98.1 \\
  
 FedAvg & 6 & 1e-4 & No & 97.6 & 93.3 & 98.8 \\
 FedAvg & 6 & 1e-4 & Yes & 0 & 93.0 & 98.2 \\
 
 FedAvg & 6 & 1e-3 & No & 97.1 & 93.0 & 98.5 \\
 FedAvg & 6 & 1e-3 & Yes & 0.5 & 91.8 & 97.6 \\
  
 FedAvg & 6 & 5e-3 & No & 93.9 & 85.4 & 91.2 \\
 FedAvg & 6 & 5e-3 & Yes & 9.2 & 82.8 & 86.8 \\
 \bottomrule
 \end{tabular}}
  \vspace{10 mm}
 \qquad
 \resizebox{\columnwidth}{!}{%
\begin{tabular}{  c c c c c c c c } 
 \toprule
 Aggregation & $M$ & $\sigma$ & RLR used? & Backdoor (\%) & Validation (\%) & Base (\%) \\
 \midrule
 FedAvg* & 0 & 0 & No & 0.4 & 93.3 & 98.7 \\
 FedAvg & 0 & 0 & No & 90.1 & 93.4 & 98.7 \\
 FedAvg & 0 & 0 & Yes & 0 & 92.9 & 97.6 \\
  
 FedAvg & 0 & 1e-4 & No & 90.7 & 93.3 & 99.0 \\
 FedAvg & 0 & 1e-4 & Yes & 0 & 92.7 & 98.2 \\
 
 FedAvg & 0 & 1e-3 & No & 90.8 & 93.3 & 98.6 \\
 FedAvg & 0 & 1e-3 & Yes & 0 & 92.5 & 97.8 \\
  
 FedAvg & 0 & 5e-3 & No & 91.3 & 93.4 & 98.9 \\
 FedAvg & 0 & 5e-3 & Yes & 0.1 & 92.6 & 97.7 \\
 
 FedAvg & 2 & 0 & No & 86.7 & 93.2 & 99.1 \\
 FedAvg & 2 & 0 & Yes & 0 & 92.7 & 98.2 \\
 
 FedAvg & 2 & 1e-4 & No & 85.1 & 93.4 & 98.8 \\
 FedAvg & 2 & 1e-4 & Yes & 0 & 92.8 & 98.1 \\
  
 FedAvg & 2 & 1e-3 & No & 87.4 & 93.5 & 99.0 \\
 FedAvg & 2 & 1e-3 & Yes & 0 & 92.5 & 97.9 \\
 
 FedAvg & 2 & 5e-3 & No & 54.8 & 91.2 & 97.7 \\
 FedAvg & 2 & 5e-3 & Yes & 1.5 & 89.8 & 97.0 \\
  
 FedAvg & 4 & 0 & No & 89.5 & 93.5 & 99.0 \\
 FedAvg & 4 & 0 & Yes & 0.1 & 93.0 & 97.7 \\
  
 FedAvg & 4 & 1e-4 & No & 88.0 & 93.4 & 98.5 \\
 FedAvg & 4 & 1e-4 & Yes & 0 & 93.2 & 97.9 \\

 FedAvg & 4 & 1e-3 & No & 84.2 & 93.3 & 98.8 \\
 FedAvg & 4 & 1e-3 & Yes & 0.3 & 92.1 & 97.4 \\
  
 FedAvg & 4 & 5e-3 & No & 50.4 & 88.5 & 94.2 \\
 FedAvg & 4 & 5e-3 & Yes & 2.6 & 87.5 & 96.3 \\
 
 FedAvg & 6 & 0 & No & 90.6 & 93.4 & 99.0 \\
 FedAvg & 6  & 0 & Yes & 0.1 & 93.2 & 98.3 \\
  
 FedAvg & 6  & 1e-4 & No & 89.4 & 93.3 & 98.4 \\
 FedAvg & 6  & 1e-4 & Yes & 0 & 92.5 & 97.3 \\
 
 FedAvg & 6  & 1e-3 & No & 91.8 & 93.0 & 97.8 \\
 FedAvg & 6  & 1e-3 & Yes & 0.1 & 92.1 & 97.7 \\
  
 FedAvg & 6  & 5e-3 & No & 20.3 & 85.2 & 91.7 \\
 FedAvg & 6  & 5e-3 & Yes & 6.1 & 84.4 & 91.2 \\
 \bottomrule
 \end{tabular}}
 \label{tab:appendixFedAvgIID}
 \end{table*}

\begin{table*}
\caption{Results for FedAvg in \niid setting under different trojans. Top-left/right: plus/square, bottom-left/right: copyright/Apple logo.}
\resizebox{\columnwidth}{!}{%
\begin{tabular}{  c c c c c c c c } 
 \toprule
 Aggregation & $M$ & $\sigma$ & RLR used? & Backdoor (\%) & Validation (\%) & Base (\%) \\
 \midrule
 FedAvg* & 0 & 0 & No & 0 & 98.5 & 99.0 \\
 FedAvg & 0 & 0 & No & 99.5 & 98.5 & 98.9 \\
 FedAvg & 0 & 0 & Yes & 0 & 95.2 & 97.7 \\
  
 FedAvg & 0 & 1e-4 & No & 99.6 & 98.5 & 99.0 \\
 FedAvg & 0 & 1e-4 & Yes & 0 & 95.3 & 98.1 \\
 
 FedAvg & 0 & 1e-3 & No & 99.6 & 98.5 & 99.0 \\
 FedAvg & 0 & 1e-3 & Yes & 0 & 95.1 & 98.3 \\
  
 FedAvg & 0 & 5e-3 & No & 99.6 & 98.5 & 99.1 \\
 FedAvg & 0 & 5e-3 & Yes & 0 & 95.1 & 98.0 \\
 
 FedAvg & 0.25 & 0 & No & 99.5 & 96.7 & 98.4 \\
 FedAvg & 0.25 & 0 & Yes & 0 & 90.0 & 97.0 \\
 
 FedAvg & 0.25 & 1e-4 & No & 99.6 & 96.7 & 98.6 \\
 FedAvg & 0.25 & 1e-4 & Yes & 0 & 89.8 & 97.7 \\
  
 FedAvg & 0.25 & 1e-3 & No & 99.6 & 96.6 & 98.3 \\
 FedAvg & 0.25 & 1e-3 & Yes & 0 & 89.9 & 96.6 \\
 
 FedAvg & 0.25 & 5e-3 & No & 99.5 & 96.6 & 98.5 \\
 FedAvg & 0.25 & 5e-3 & Yes & 0 & 89.8 & 96.1 \\
  
 FedAvg & 0.5 & 0 & No & 99.5 & 98.0 & 98.8 \\
 FedAvg & 0.5 & 0 & Yes & 0 & 93.8 & 97.9 \\
  
 FedAvg & 0.5 & 1e-4 & No & 99.5 & 98.1 & 98.8 \\
 FedAvg & 0.5 & 1e-4 & Yes & 0 & 93.7 & 97.18 \\

 FedAvg & 0.5 & 1e-3 & No & 99.5 & 98.1 & 98.7 \\
 FedAvg & 0.5 & 1e-3 & Yes & 0 & 93.8 & 97.5 \\
  
 FedAvg & 0.5 & 5e-3 & No & 99.4 & 97.7 & 98.4 \\
 FedAvg & 0.5 & 5e-3 & Yes & 0 & 94.4 & 97.4 \\
 
 FedAvg & 1.0 & 0 & No & 99.6 & 98.4 & 99.0 \\
 FedAvg & 1.0 & 0 & Yes & 0 & 94.9 & 97.8 \\
  
 FedAvg & 1.0 & 1e-4 & No & 99.4 & 98.5 & 98.8 \\
 FedAvg & 1.0 & 1e-4 & Yes & 0 & 94.8 & 97.8 \\
 
 FedAvg & 1.0 & 1e-3 & No & 99.6 & 98.3 & 98.9 \\
 FedAvg & 1.0 & 1e-3 & Yes & 0 & 94.6 & 97.4 \\
  
 FedAvg & 1.0 & 5e-3 & No & 99.2 & 98.5 & 99.1 \\
 FedAvg & 1.0 & 5e-3 & Yes & 28.4 & 94.9 & 97.5 \\
 \bottomrule
 \end{tabular}}
 \vspace{10 mm}
 \qquad
\resizebox{\columnwidth}{!}{%
\begin{tabular}{  c c c c c c c c } 
 \toprule
 Aggregation & $M$ & $\sigma$ & RLR used? & Backdoor (\%) & Validation (\%) & Base (\%) \\ 
 \midrule
 FedAvg* & 0 & 0 & No & 0.7 & 93.2 & 99.0 \\
 FedAvg & 0 & 0 & No & 95.0 & 93.3 & 98.5 \\
 FedAvg & 0 & 0 & Yes & 0 & 92.6 & 98.5 \\
  
 FedAvg & 0 & 1e-4 & No & 94.9 & 93.4 & 98.3 \\
 FedAvg & 0 & 1e-4 & Yes & 0 & 92.4 & 98.3 \\
 
 FedAvg & 0 & 1e-3 & No & 95.9 & 93.4 & 98.8 \\
 FedAvg & 0 & 1e-3 & Yes & 0 & 92.5 & 97.9 \\
  
 FedAvg & 0 & 5e-3 & No & 94.0 & 93.6 & 99.0 \\
 FedAvg & 0 & 5e-3 & Yes & 0 & 92.6 & 98.3 \\
 
 FedAvg & 0.25 & 0 & No & 96.5 & 93.4 & 98.9 \\
 FedAvg & 0.25 & 0 & Yes & 0 & 92.7 & 98.4 \\
 
 FedAvg & 0.5 & 1e-4 & No & 94.2 & 93.4 & 98.7 \\
 FedAvg & 0.5 & 1e-4 & Yes & 0 & 92.8 & 98.1 \\
  
 FedAvg & 0.25 & 1e-3 & No & 93.8 & 93.6 & 99.3 \\
 FedAvg & 0.25 & 1e-3 & Yes & 0 & 92.6 & 98.0 \\
 
 FedAvg & 0.25 & 5e-3 & No & 42.6 & 91.2 & 97.5 \\
 FedAvg & 0.25 & 5e-3 & Yes & 1.4 & 89.8 & 96.3 \\
  
 FedAvg & 0.5 & 0 & No & 95.0 & 93.5 & 98.8 \\
 FedAvg & 0.5 & 0 & Yes & 0.1 & 93.2 & 98.3 \\
  
 FedAvg & 0.5 & 1e-4 & No & 93.7 & 93.5 & 99.0 \\
 FedAvg & 0.5 & 1e-4 & Yes & 0 & 93.4 & 97.7 \\
 
 FedAvg & 0.5 & 1e-3 & No & 94 & 93.4 & 99.0 \\
 FedAvg & 0.5 & 1e-3 & Yes & 0.5 & 92.3 & 97.5 \\
  
 FedAvg & 0.5& 5e-3 & No & 33.6 & 88.8 & 97.1 \\
 FedAvg & 0.5 & 5e-3 & Yes & 3.1 & 87.2 & 94.0 \\
 
 FedAvg & 1 & 0 & No & 93.2 & 93.4 & 98.5 \\
 FedAvg & 1 & 0 & Yes & 0 & 93.1 & 97.7 \\
  
 FedAvg & 1 & 1e-4 & No & 94 & 93.4 & 99.2 \\
 FedAvg & 1 & 1e-4 & Yes & 0 & 93.1 & 98.5 \\
 
 FedAvg & 1 & 1e-3 & No & 94.3 & 92.9 & 98.5 \\
 FedAvg & 1 & 1e-3 & Yes & 0.5 & 91.4 & 98.1 \\
  
 FedAvg & 1 & 5e-3 & No & 25.4 & 85.3 & 94.5 \\
 FedAvg & 1 & 5e-3 & Yes & 6.1 & 83.3 & 88.2 \\
 \bottomrule
 \end{tabular}}
  \vspace{10 mm}
\resizebox{\columnwidth}{!}{%
\begin{tabular}{  c c c c c c c c } 
 \toprule
 Aggregation & $M$ & $\sigma$ & RLR used? & Backdoor (\%) & Validation (\%) & Base (\%) \\
 \midrule
 FedAvg* & 0 & 0 & No & 0.1 & 98.5  & 99.1 \\
 FedAvg & 0 & 0 & No & 99.5 & 98.5 & 99.0 \\
 FedAvg & 0 & 0 & Yes & 0 & 94.9 & 97.6 \\
  
 FedAvg & 0 & 1e-4 & No & 99.5 & 98.5 & 99.0 \\
 FedAvg & 0 & 1e-4 & Yes & 0 & 95.1 & 97.8 \\
 
 FedAvg & 0 & 1e-3 & No & 99.5 & 98.5 & 99.1 \\
 FedAvg & 0 & 1e-3 & Yes & 0 & 95.2 & 97.7 \\
  
 FedAvg & 0 & 5e-3 & No & 99.6 & 98.5 & 99.0 \\
 FedAvg & 0 & 5e-3 & Yes & 0 & 95.2 & 98.1 \\
 
 FedAvg & 0.25 & 0 & No & 99.4 & 96.6 & 98.5 \\
 FedAvg & 0.25 & 0 & Yes & 0 & 89.3 & 97.1 \\
 
 FedAvg & 0.25 & 1e-4 & No & 99.4 & 96.9 & 98.6 \\
 FedAvg & 0.25 & 1e-4 & Yes & 0 & 89.5 & 97.0 \\
  
 FedAvg & 2 & 1e-3 & No & 99.3 & 96.9 & 98.5 \\
 FedAvg & 2 & 1e-3 & Yes & 0 & 89.2 & 96.3 \\
 
 FedAvg & 0.25 & 5e-3 & No & 99.2 & 96.7 & 98.4 \\
 FedAvg & 0.25 & 5e-3 & Yes & 0 & 90.2 & 96.4 \\
  
 FedAvg & 0.5 & 0 & No & 99.4 & 98.0 & 98.9 \\
 FedAvg & 0.5 & 0 & Yes & 0.0 & 93.2 & 96.9 \\
  
 FedAvg & 0.5 & 1e-4 & No & 99.5 & 98.0 & 98.8 \\
 FedAvg & 0.5 & 1e-4 & Yes & 0 & 93.5 & 97.7 \\
 
 FedAvg & 0.5 & 1e-3 & No & 99.4 & 98.1 & 98.9 \\
 FedAvg & 0.5 & 1e-3 & Yes & 0 & 93.4 & 96.9 \\
  
 FedAvg & 0.5 & 5e-3 & No & 99.5 & 97.8 & 98.5 \\
 FedAvg & 0.5 & 5e-3 & Yes & 0 & 94.3 & 97.6 \\
 
 FedAvg & 1 & 0 & No & 99.5 & 98.5 & 99.0 \\
 FedAvg & 1 & 0 & Yes & 0 & 95.1 & 98.0 \\
  
 FedAvg & 1 & 1e-4 & No & 99.6 & 98.4 & 99.1 \\
 FedAvg & 1 & 1e-4 & Yes & 0 & 95.0 & 98.1 \\
 
 FedAvg & 1 & 1e-3 & No & 99.5 & 98.4 & 98.9 \\
 FedAvg & 1 & 1e-3 & Yes & 0 & 95.0 & 98.0 \\
  
 FedAvg & 1 & 5e-3 & No & 99.4 & 98.0 & 98.7 \\
 FedAvg & 1 & 5e-3 & Yes & 0 & 94.9 & 97.7 \\
 \bottomrule
 \end{tabular}}
  \vspace{10 mm}
 \qquad
 \resizebox{\columnwidth}{!}{%
\begin{tabular}{  c c c c c c c c } 
 \toprule
 Aggregation & $M$ & $\sigma$ & RLR used? & Backdoor (\%) & Validation (\%) & Base (\%) \\ 
 \midrule
  FedAvg* & 0 & 0 & No & 0 & 98.5  & 99.1 \\
 FedAvg & 0 & 0 & No & 99.4 & 98.5 & 99.1 \\
 FedAvg & 0 & 0 & Yes & 0 & 95.0 & 97.6 \\
  
 FedAvg & 0 & 1e-4 & No & 99.6 & 98.5 & 99.0 \\
 FedAvg & 0 & 1e-4 & Yes & 0 & 95.4 & 98.1 \\
 
 FedAvg & 0 & 1e-3 & No & 99.5 & 98.4 & 99.0 \\
 FedAvg & 0 & 1e-3 & Yes & 0 & 95.0 & 97.7 \\
  
 FedAvg & 0 & 5e-3 & No & 99.3 & 98.5 & 99.0 \\
 FedAvg & 0 & 5e-3 & Yes & 0 & 95.2 & 97.6 \\
 
 FedAvg & 0.25 & 0 & No & 99.5 & 96.8 & 98.6 \\
 FedAvg & 0.25 & 0 & Yes & 0 & 89.2 & 97.4 \\
 
 FedAvg & 0.25 & 1e-4 & No & 99.5 & 96.5 & 98.6 \\
 FedAvg & 0.25 & 1e-4 & Yes & 0 & 89.3 & 97.0 \\
  
 FedAvg & 2 & 1e-3 & No & 99.4 & 97.0 & 98.5 \\
 FedAvg & 2 & 1e-3 & Yes & 0 & 89.8 & 96.4 \\
 
 FedAvg & 0.25 & 5e-3 & No & 99.2 & 96.4 & 98.6 \\
 FedAvg & 0.25 & 5e-3 & Yes & 0 & 90.7 & 96.7 \\
  
 FedAvg & 0.5 & 0 & No & 99.5 & 98.0 & 98.6 \\
 FedAvg & 0.5 & 0 & Yes & 0.0 & 93.4 & 98.2 \\
  
 FedAvg & 0.5 & 1e-4 & No & 99.5 & 98.0 & 98.8 \\
 FedAvg & 0.5 & 1e-4 & Yes & 0 & 93.5 & 97.7 \\
 
 FedAvg & 0.5 & 1e-3 & No & 99.4 & 98.0 & 98.9 \\
 FedAvg & 0.5 & 1e-3 & Yes & 0 & 93.5 & 96.9 \\
  
 FedAvg & 0.5 & 5e-3 & No & 99.3 & 97.8 & 98.6 \\
 FedAvg & 0.5 & 5e-3 & Yes & 0 & 94.2 & 97.2 \\
 
 FedAvg & 1 & 0 & No & 99.5 & 98.5 & 99.0 \\
 FedAvg & 1 & 0 & Yes & 0 & 95.1 & 98.0 \\
  
 FedAvg & 1 & 1e-4 & No & 99.6 & 98.4 & 99.0 \\
 FedAvg & 1 & 1e-4 & Yes & 0 & 94.9 & 97.8 \\
 
 FedAvg & 1 & 1e-3 & No & 99.5 & 98.4 & 98.7 \\
 FedAvg & 1 & 1e-3 & Yes & 0 & 94.9 & 97.9 \\
  
 FedAvg & 1 & 5e-3 & No & 99.5 & 97.9 & 98.6 \\
 FedAvg & 1 & 5e-3 & Yes & 0 & 94.8 & 98.0 \\
 \bottomrule
 \end{tabular}}
 \label{tab:appendixFedAvgNIID}
 \end{table*}
 
\clearpage
\section{Proof of Convergence Rate}
\label{app:proof}

\subsection{Preliminaries}
Please note that $\eta_{\theta, i}$ defined in the main text could be defined using an indicator $I$ function. We choose this notation to simplify the steps in the proof.
\begin{eqnarray}
\label{majority}
\eta_{\theta, i} &=&  
\begin{cases}
\eta & \left | \sum_{k \in S_t} \sign(\Delta_{t,i}^k) \right | \geq \theta \\
-\eta  & \text{otherwise}
\end{cases} \\
 &=&\eta.(2 \cdot \IND -1)
\label{eqn:rlrmatrix}
\end{eqnarray}
In order to represent the component wise multiplication of the learning rate with update discussed in Section~\ref{sec:robustLR} as matrix multiplication, we represent 
$\IND$ as $d \times d$ matrix $I_t$ where $d$ is the size of $w_t$, and $\forall i,j \in [1..d],$
$I_t[i,i]= \IND $ and $I_t[i,j]=0$ where $i \neq j$. 
This implies that our update rule can be written as (assuming equal weight for all $n$ participants , and $\mathbb{I}$ is a $d \times d$ identity matrix.)
\[w_{t+1} = w_t + 
\eta \cdot(2.I_t- \mathbb{I}) 
 \left( \frac{1}{n}{\sum_{k=1}^n \Delta_t^k} \right )\]
Furthermore, assuming each party sends its update after single iteration of local stochastic gradient descent (SGD), then we get $\Delta_t^k= - \nabla f_k(w_t)$ as the update sent by each party. 
Hence, using matrix notation,  after single local SGD updates, 
our model update is equal to
\[w_{t+1} = w_t - 
\eta \cdot(2.I_t- \mathbb{I}) 
\left( \frac{1}{n}{\sum_{k=1}^n \nabla f_k(w_t)} \right)\]

\subsection{Assumptions}

\medskip
{\bf Proof} Note that given the proposed supermajority rule~(\ref{eqn:rlrmatrix}), $||2I_t-\mathbb{I}||_2=1$, where $||\cdot||_2$ is an operator norm. 
In view of our update rule~(\ref{eqn:rlrmatrix}), we have
{\footnotesize
\begin{eqnarray}
\label{main}
f(\hat{w}_{t+1}) &\leq&  f({\hat{w}_t})+<\nabla f(\hat{w}_t), \hat{w}_{t+1}-\hat{w}_t>\\
 && + \frac{L}{2} ||\hat{w}_{t+1}-\hat{w}_t||^2 \nonumber \\ &=&f(\hat{w}_t)+ \frac{L\eta^2}{2} \biggl \|(2I_t-\mathbb{I}) \frac{1}{n} \sum_{k=1}^n \nabla f_k(w^k_{t-1}, \xi^{t+1}_k)\biggr\|^2  \nonumber \\
&&  - \eta <\nabla f(\hat{w}_t), (2I_t-\mathbb{I}) \frac{1}{n} \sum_{k=1}^n \nabla f_k(w^k_{t}, \xi^{t+1}_k)> \nonumber\\
&&  + \frac{L\eta^2}{2} \biggl \|(2I_t-\mathbb{I}) \frac{1}{n} \sum_{k=1}^n \nabla f_k(w^k_{t-1}, \xi^{t+1}_k)\biggr\|^2. \nonumber 
\end{eqnarray}
}

Now account that $||2I_t-\mathbb{I}||_2=1$
and
take conditional expectation in respect to filtration $\digamma_t$ generated by all random variables at step $t$, i.e. a sequence of increasing $\sigma$-algebras $\digamma_s \subseteq \digamma_t$ for all $s<t$:
{\footnotesize
\begin{eqnarray}
\label{exp_main}
&& \mathbb{E}_{\digamma_t} f(\hat{w}_{t+1}) \leq  f(\hat{w}_t) + \frac{L\eta^2}{2} \mathbb{E}_{\digamma_t}\biggl \|\frac{1}{n} \sum_{k=1}^n \nabla f_k(w^k_{t}, \xi^{t+1}_k)\biggr \|^2 \nonumber \\ 
&&  -  \eta<\nabla f(\hat{w}_t), \mathbb{E}_{\digamma_t}\biggl\{   \frac{2I_t-\mathbb{I}}{n} \sum_{k=1}^n \nabla f_k(w^k_{t}, \xi^{t+1}_k)\biggr\}>. 
\end{eqnarray}
}

Note that a case of $Pr(I_t[i,i]=1|\digamma_t)$ for all $i=1, \ldots, d$ corresponds to a standard federated learning framework (i.e., $\IND =1$).
To avoid burdening the reader with cumbersome element-wise matrix derivations, we derive a convergence result under a scenario that for a given $t\in \mathbb{Z}$ we change either signs of none or all elements, i.e. $I_t[i,i]\equiv 0$ or $I_t[i,i]\equiv 0$ for $i=1, \ldots, d$, respectively.   For the sake of notation we refer to these cases as $I_t=1|\digamma_t$ and $I_t=0|\digamma_t$. Let $p_{I_t=0|\digamma_t}$ and $p_{I_t=1|\digamma_t}$) be respectively  be probability that $I_t=0|\digamma_t$ and  $I_t=1|\digamma_t$.  The case when for a given $t\in \mathbb{Z}$ 
there exist indices $i,j=1, \ldots, d$ such that
$I_t[i,i]=0$ and $I_t[j,j]=0$ is addressed verbatim but is more tedious. Nevertheless, all necessary assumptions are stated under the element-wise case. {
Under~(\ref{pop_symm}) 
the sign of a random variable is independent of its absolute value~\cite{wolfe1974characterization}.} Hence,
\footnotesize{
\begin{eqnarray}
\label{mid}
&&\mathbb{E}_{\digamma_t}\biggl\{  \frac{2I_t-\mathbb{I}}{n} \sum_{k=1}^n \nabla f_k(w^k_{t}, \xi^{t+1}_k)\biggr\} \\ &&=
-\frac{1}{n} \sum_{k=1}^n \nabla f_k(w^k_{t})p_{I_t=0|\digamma_t} +\frac{1}{n} \sum_{k=1}^n \nabla f_k(w^k_{t})p_{I_t=1|\digamma_t} \nonumber \\ &&= \frac{1}{n} \sum_{k=1}^n \nabla f_k(w^k_{t})(1-2p_{I_t=0|\digamma_t}). \nonumber
\end{eqnarray}
}

In view of the Cauchy-Schwartz inequality and Assumption~2, the last term in~(\ref{exp_main}) $T4= \mathbb{E}_{\digamma_t}\biggl \|\frac{1}{n} \sum_{k=1}^n \nabla f_k(w^k_{t}, \xi^{t+1}_k)\biggr \|^2$ can be rewritten as
\footnotesize{
\[ \mathbb{E}_{\digamma_t}\biggl \|\frac{1}{n} \sum_{k=1}^n \nabla f_k(w^k_{t}, \xi^{t+1}_k)-\frac{1}{n}\sum_{k=1}^n \nabla f_k(w^k_t)+ \frac{1}{n}\sum_{k=1}^n \nabla f_k (w^k_t)\biggr \|^2 \nonumber \\\]
}
 and $T4$ can be bounded as
\footnotesize{
\begin{eqnarray}
\label{M3}
T4 &\leq&
\mathbb{E}_{\digamma_t}\biggl \|\frac{1}{n} \sum_{k=1}^n \nabla f_k(w^k_{t}, \xi^{t+1}_k)-\frac{1}{n}\sum_{k=1}^n \nabla f_k(w^k_t)\biggr \|^2 \nonumber\\
&&+\mathbb{E}_{\digamma_t}\biggl \|\frac{1}{n}\sum_{k=1}^n \nabla f_k(w^k_t)\biggr \|^2 \\ &\leq& \frac{\sigma^2}{n}+\biggl \|\frac{1}{n}\sum_{k=1}^n \nabla f_k(w^k_t)\biggr \|^2.
\end{eqnarray}
}

Hence, plugging in~(\ref{M3}) and~(\ref{mid}) into~(\ref{exp_main}) 
and
taking expectation over all random variables in~(\ref{exp_main}) yields
\footnotesize{
\begin{eqnarray}
\label{interm}
\mathbb{E}f(\hat{w}_{t+1}) && \leq
\mathbb{E}f(\hat{w}_{t}) -\frac{\eta}{2}
\mathbb{E}||\nabla f(\hat{w}_t)||^2 \\
&& -\frac{\eta}{2} \mathbb{E}\biggl \|(1-2p_{I_t=0|\digamma_t})\frac{1}{n}\sum_{k=1}^n \nabla f_k(w^k_t)\biggr \|^2 \nonumber \\ &&
+ \frac{\eta}{2} \mathbb{E}\biggl \|\nabla f(\hat{w}_t)-(1-2p_{I_t=0|\digamma_t})\frac{1}{n}\sum_{k=1}^n \nabla f_k(w^k_t)\biggr \|^2  \nonumber \\
&& +\frac{L\eta^2}{2}\mathbb{E}\biggl \|\frac{1}{n}\sum_{k=1}^n \nabla f_k(w^k_t)\biggr \|^2
+\frac{L\eta^2\sigma^2}{2n}. \nonumber
\end{eqnarray}
}

Consider the forth term in~(\ref{interm})
\footnotesize{
\begin{eqnarray}
\label{diff}
&&\mathbb{E}\biggl \|\nabla f(\hat{w}_t)-(1-2p_{I_t=0|\digamma_t})\frac{1}{n}\sum_{k=1}^n \nabla f_k(w^k_t)\biggr \|^2  \\ && \leq
 \mathbb{E}||2p_{I_t=0|\digamma_t} \nabla f(\hat{w}_t)||^2 
+
\mathbb{E}\biggl \|\nabla f_k(w^k_t)-\frac{1}{n}\sum_{k=1}^n \nabla f_k(\hat{w}_t)\biggr \|^2 \nonumber \\
&&\leq  \mathbb{E}||2p_{I_t=0|\digamma_t} \nabla f(\hat{w}_t)||^2
+L^2 \mathbb{E}||\frac{1}{n}\sum_{k=1}^n (\hat{w}_t-w^k_t)||^2 \nonumber \\
&&\leq  \mathbb{E}||2p_{I_t=0|\digamma_t} \nabla f(\hat{w}_t)||^2
+L^2M^2,
\end{eqnarray}
}
where we account for unit norm of $2I_t-\mathbb{I}$ and Assumption~1 on smoothness of each $\nabla f_k$ with modulus $L$.

Plugging in~(\ref{diff}) into~(\ref{interm}) 
yields
\footnotesize{
\begin{eqnarray}
\label{almost}
&&\mathbb{E}f(\hat{w}_{t+1})\leq 
\mathbb{E}f(\hat{w}_{t}) -\frac{\eta}{2}
\mathbb{E}||\nabla f(\hat{w}_t)||^2 \\
&&- \frac{\eta}{2}(1-4p_{I_t=0|\digamma_t}-L\nu)\mathbb{E}\biggl \|\frac{1}{n}\sum_{k=1}^n \nabla f_k(w^k_t)\biggr \|^2 \nonumber \\
&&+\frac{L^2 M^2\eta}{2}+\frac{L\eta^2\sigma^2}{2n}. \nonumber
\end{eqnarray}
}

Now by accounting for
$Pr(I_t[1,1]=\ldots=I_t[d,d]=0)|\digamma_t)\leq
Pr(I_t[i,i]=0|\digamma_t)\leq p_0 <0.25$, $1 \leq i\leq d$ and $0<\nu\leq (1-p_0)/L$, we find
that
\begin{eqnarray}
\label{almost}
\mathbb{E}f(\hat{w}_{t+1})\leq  
\mathbb{E}f(\hat{w}_{t})-\frac{\eta}{2}
\mathbb{E}||\nabla f(\hat{w}_t)||^2+\frac{L^2 M^2\eta}{2}+\frac{L\eta^2\sigma^2}{2n}.
\end{eqnarray}

Now telescoping~(\ref{almost}) over $t=0,\ldots, T-1$ and dividing both sides by $T$ yields
\begin{eqnarray}
\frac{1}{T}\sum_{t=0}^{T-1}
\mathbb{E}||\nabla f(\hat{w}_t)||^2
\leq \frac{2}{\eta T} (f(\hat{w}_0)-f^{*})
+ L^2M^2 +\frac{L\eta\sigma^2}{n},
\end{eqnarray}
which concludes the proof.

\medskip
\textbf{Remark: Attack Success Rate:}
Our robust learning rate scheme tries to maximize the dimensions where there is no consensus (e.g., when for the $i^{\mbox{th}}$ dimension update $\IND$ is zero.) In one sense, this could be considered a case where our algorithm believes there is some potential attack may be happening and tries to go to the reverse direction. In our convergence proof, we assume that such scenarios happen with probability less than $0.25$ of the time (i.e.,$\forall i$, $0\leq Pr(1- \IND |\digamma_t)\leq p_0 <0.25$). In other words, we go against the direction provided by the gradients in standard federated learning less than the $0.25$ probability. This assumption is reasonable in settings where attacker control only a minority of the overall agents and honest agents agree on the update direction often enough.

\end{document}